\documentclass[acmsmall]{acmart}

\usepackage{caption}
\usepackage{subcaption}
\usepackage{multirow}
\usepackage{makecell}

\AtBeginDocument{%
  \providecommand\BibTeX{{%
    \normalfont B\kern-0.5em{\scshape i\kern-0.25em b}\kern-0.8em\TeX}}}

\setcopyright{acmcopyright}
\copyrightyear{2024}
\acmYear{2024}

\acmJournal{JACM}
\acmVolume{37}
\acmNumber{4}
\acmArticle{111}
\acmMonth{8}

\begin{document}

\title{A Unified Review of Deep Learning for Automated Medical Coding}

\author{Shaoxiong Ji}
\orcid{0000-0003-3281-8002}
\email{shaoxiong.ji@helsinki.fi}
\affiliation{%
  \institution{Aalto University and University of Helsinki}
  \streetaddress{PO BOX 11000}
  \city{Espoo}
  \country{Finland}
  \postcode{02150}
}

\author{Xiaobo Li}
\affiliation{
  \institution{Dalian Maritime University}
  \city{Dalian}
  \country{China}
  \postcode{116000}
}
\email{xiaobo.li@dlmu.edu.cn}

\author{Wei Sun}
\affiliation{%
  \institution{KU Leuven}
  \streetaddress{PO BOX 11000}
  \city{Leuven}
  \country{Belgium}
  \postcode{}
}
\email{wei.sun@kuleuven.be}

\author{Hang Dong}
\affiliation{%
\institution{University of Exeter}
\city{Exeter}
 \country{UK}
}
 \email{h.dong2@exeter.ac.uk}

\author{Ara Taalas}
\affiliation{
  \institution{Institute for Molecular Medicine Finland (FIMM), HiLIFE, University of Helsinki and Terveystalo Healthcare Services}
  \country{Finland}
}
\email{ara.taalas@terveystalo.com}

\author{Yijia Zhang}
\affiliation{
  \institution{Dalian Maritime University}
  \city{Dalian}
  \country{China}
  \postcode{116000}
}
\email{zhangyijia@dlmu.edu.cn}

\author{Honghan Wu}
\affiliation{%
  \institution{School of Health and Wellbeing, University of Glasgow}
  \country{UK}
}
\email{honghan.wu@ucl.ac.uk}

\author{Esa Pitkänen}
\affiliation{
  \institution{Institute for Molecular Medicine Finland (FIMM), HiLIFE, University of Helsinki}
  \country{Finland}
}
\email{esa.pitkanen@helsinki.fi}

\author{Pekka Marttinen}
\affiliation{%
  \institution{Aalto University}
  \streetaddress{PO BOX 11000}
  \city{Espoo}
  \country{Finland}
  \postcode{02150}
}
\email{pekka.marttinen@aalto.fi}

\renewcommand{\shortauthors}{Ji, et al.}

\begin{abstract}
Automated medical coding, an essential task for healthcare operation and delivery, makes unstructured data manageable by predicting medical codes from clinical documents. 
Recent advances in deep learning and natural language processing have been widely applied to this task. 
However, deep learning-based medical coding lacks a unified view of the design of neural network architectures.
This review proposes a unified framework to provide a general understanding of the building blocks of medical coding models and summarizes recent advanced models under the proposed framework. 
Our unified framework decomposes medical coding into four main components, i.e., encoder modules for text feature extraction, mechanisms for building deep encoder architectures, decoder modules for transforming hidden representations into medical codes, and the usage of auxiliary information. 
Finally, we introduce the benchmarks and real-world usage and discuss key research challenges and future directions. 
\end{abstract}

\begin{CCSXML}
<ccs2012>
   <concept>
       <concept_id>10010405.10010444.10010447</concept_id>
       <concept_desc>Applied computing~Health care information systems</concept_desc>
       <concept_significance>500</concept_significance>
       </concept>
   <concept>
       <concept_id>10010147.10010178.10010179</concept_id>
       <concept_desc>Computing methodologies~Natural language processing</concept_desc>
       <concept_significance>300</concept_significance>
       </concept>
   <concept>
       <concept_id>10010405.10010497</concept_id>
       <concept_desc>Applied computing~Document management and text processing</concept_desc>
       <concept_significance>100</concept_significance>
       </concept>
 </ccs2012>
\end{CCSXML}

\ccsdesc[500]{Applied computing~Health care information systems}
\ccsdesc[300]{Computing methodologies~Natural language processing}
\ccsdesc[100]{Applied computing~Document management and text processing}

\keywords{Medical Coding, Deep Learning, Unified Framework}

\maketitle

\section{Introduction}
\label{sec:introduction}
In the field of Natural Language Processing (NLP), deep learning that builds deep neural networks for representation learning has attracted significant attention from the research community and achieved superior performance in various applications such as automatic extraction of useful information and answer generation from human queries~\cite{otter2020survey,torfi2020natural}. 
\textcolor{black}{
NLP techniques allow the machine to process human languages automatically and have been widely studied in analyzing health-related texts, measuring healthcare quality, and promoting the delivery of healthcare services.
For example, Sentic PROMs (patient-reported outcome measures) enable patients' physio-emotional sensitivity tracking and measuring healthcare quality through sentic computing on free-text patient notes~\cite{cambria2012sentic}.
Contextualized text representations and classification models also facilitate outbreak management during epidemics~\cite{khatua2019tale}.
There are other patient-centered applications, such as proactive mental healthcare~\cite{ji2022suicidal} and patient opinion mining~\cite{cambria2010sentic}, to name a few.
}
This review focuses on deep neural network-based NLP techniques for automated medical coding with medical ontologies, also known as medical code assignment, medical code prediction, medical coding, or clinical coding.
Medical code assignment uses all types of clinical notes to predict medical codes in a supervised manner with human-annotated codes~\cite{perotte2014diagnosis}, formulated as a multi-class multi-label text classification problem in the medical domain.
\textcolor{black}{
Most deep learning-based medical coding models are trained in a centralized manner, while some recent publications investigate the emerging application of federated learning~\cite{chen2022training}.
}

Healthcare workers write clinical notes about a patient's health status to document their insights and observations for further diagnosis decision support. 
Clinical notes as free-text descriptions are an essential component of Electronic Health Records (EHRs), which contain patient medical history, symptom description, lab test result summary, reasons for diagnoses, and daily activities~\cite{li2022neural}.
Diagnosis codes in typical medical classification systems identify a patient's diseases, disorders, symptoms, and specific reasons for the hospital visit. 
In contrast, procedural codes or intervention codes identify surgical, medical, or diagnostic interventions.  
Diagnosis codes, which a trained health professional assigns, act as the standard translation of written patient descriptions. 
Diagnostic coding is an integral part of the clinical coding process in health information management with procedural codes.
\textcolor{black}{
Medical coding, particularly for billing purposes, often operates separately from the clinical care process and may not significantly impact immediate patient care decisions.
While medical coding primarily serves billing and administrative purposes, accurate coding does have broader implications for healthcare quality assessment, research, and resource allocation~\cite{burns2012systematic,nair2013ensuring,burks2022systematic}. 
It can play a role in retrospective analysis, identifying trends in patient populations, and assessing the effectiveness of certain treatments or interventions.
}

\begin{figure}[htbp]
\centering
\includegraphics[width=0.7\linewidth]{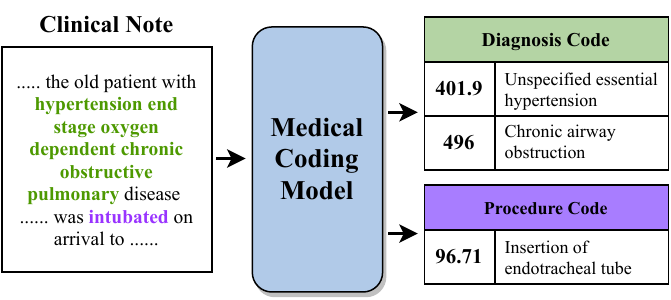}
\caption{A medical coding model maps an example clinical note to the corresponding ICD procedure and diagnosis codes.}
\label{fig:example}
\end{figure}

Clinical notes are usually annotated with standardized statistical codes to facilitate information management. 
Different diagnosis classification systems utilize various medical coding systems.  
The International Classification of Diseases (ICD) system, maintained by the World Health Organization (WHO), is one of the most widely-used coding systems adopted in countries across the globe\footnote{\url{https://www.who.int/standards/classifications/classification-of-diseases}}.
The ICD system transforms diseases, symptoms, signs, and treatment procedures into standard medical codes. 
\textcolor{black}{
It has been widely used for clinical data analysis, automated medical decision support~\cite{choi2016doctor}, billing, and medical insurance reimbursement~\cite{park2000accuracy}. 
}
Specific versions of ICD include ICD-9, ICD-9-CM, ICD-10, and ICD-11. 
Most ICD-9 codes consist of three digits to the left of a decimal point and one or two digits to the right.
Some ICD-9 codes have ``V'' or ``E'' in front of the digits, representing preventive health services and environmental causes of health problems.
Figure~\ref{fig:example} shows a fragment of the patient's clinical note with ICD-9-CM codes assigned. 
The ICD-9-CM created by the US National Center for Health Statistics (NCHS) adapts ICD-9 codes used in the United States.
The first three characters of ICD-10 codes define the category, and the next three digits describe the etiology, anatomic site, severity, and other vital information. 
The latest ICD version is ICD-11, which will become effective in 2022, while older versions such as ICD-9, ICD-9-CM, and ICD-10 are also concurrently used. 
\textcolor{black}{
Other widely used medical condition classification systems include the Clinical Classifications Software (CCS) and Hierarchical Condition Category (HCC) coding. It is worth noting that CCS and HCC are derived directly from ICD codes and serve as specialized, broader categorizations within the ICD framework, tailored for specific purposes.
}

\textcolor{black}{Properly coded medical information is vital for clinical decision-making, public health surveillance, research, and reimbursement.} 
Automated care pathways are often triggered by patients receiving a specific diagnosis code. On the national scale, care guidelines are often structured around diagnosis codes, providing interventions for clearly defined conditions~\cite{currentcareguidelines, wong2020shaping}. On the healthcare provider side, quantitative measurement of healthcare effectiveness and care development is, by necessity, based on code-based logic. Questions, such as how many patients diagnosed with a given illness received appropriate care, can only be measured quantitatively based on medical coding~\cite{donaldson1999measuring}. Automated diagnosis coding can also be deployed to detect missed diagnoses and adverse effects~\cite{melton2005automated}. Effective treatment often relies on the early detection of symptoms, and pre-emptive healthcare can only be built on technology sensitive to slight deviations in the patient's health state.

Automatic medical code assignment uses feature engineering techniques and machine learning-based classifiers to predict medical codes from clinical notes~\cite{crammer2007automatic}.
Medical coding requires efficient matching between textual mentions and specific diagnosed codes. 
It exploits the dependencies between input and output variables by learning structured output representation.
Traditional medical coding systems deploy rule-based methods~\cite{farkas2008automatic}, select manual features~\cite{medori2010machine}, and apply machine learning-based classification models such as Support Vector Machines (SVM)~\cite{boytcheva2011automatic} and Bayesian ridge regression~\cite{lita2008large}. 
The hierarchical structure of the code system, depicted as a tree structure with multiple levels, is a basic pattern for improving the automated coding method. 
For example, Perotte et al.~\cite{perotte2014diagnosis} adopted the ICD hierarchy and developed flat and hierarchical SVM for diagnosis code classification.
The breakthrough of natural language processing with deep neural networks has led to neural classifiers with word embedding and deep learning~\cite{wang2020using}.
Neural methods for medical text encoding intensively use recurrent neural networks (RNNs), convolutional neural networks (CNNs), and neural attention mechanisms, where parameter selection is a vital issue~\cite{liu2021parameter}.

\begin{figure}[htbp!]
\begin{center}
\includegraphics[width=0.5\linewidth]{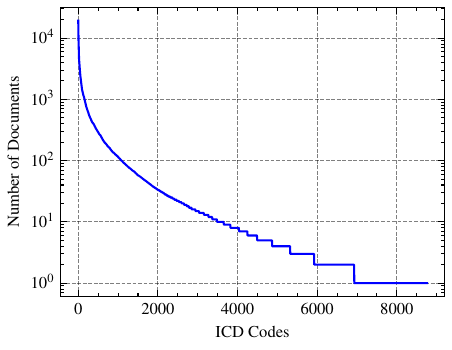}
\caption{\textcolor{black}{The distribution of ICD codes in the MIMIC-III dataset curated by \citet{mullenbach2018explainable}}.}
\label{fig:ICD_dist}
\end{center}
\end{figure}

However, three main challenges remain in processing medical text and automated coding.
\paragraph{(1) Noisy and Lengthy Clinical Notes.}
Clinical notes contain many professional medical vocabularies and noisy information such as non-standard synonyms and misspellings.
They are usually lengthy documents containing many types of clinical information, such as health profiles, lab tests, radiology reports, operative reports, and medications.
Thus, they typically have hundreds or even thousands of words. 
Some patients with long hospital stays may have much longer written notes. \textcolor{black}{Additionally, writing styles can vary from one healthcare professional to another, with domain-specific lingo giving a given word different meanings depending on context. The medical practice also evolves with time, with coding systems and notation changing from one year to another.}
\paragraph{(2) High-dimensional Medical Codes.}
Medical notes are associated with multiple diagnoses, usually treated as a multi-label extreme classification problem containing a large label set.
The high-dimensional label space has thousands of codes. For example, ICD-9 and ICD-10 coding systems have more than 14,000 and 68,000 codes, respectively. 
The space of target classes is exponential to the number of output classes making it extremely challenging when facing high-dimensional medical ontologies.
\paragraph{(3) Imbalanced Classes}
A patient typically is diagnosed with only a couple of codes over the whole coding space, while patients with complicated diseases are associated with dozens of codes.
Moreover, because of the existence of common and rare diseases, the distribution of medical codes in an EHR system is imbalanced, also known as the long-tail phenomenon. 
For example, the distribution is highly skewed in the MIMIC-III dataset~\cite{johnson2016mimic}, as shown in Fig.~\ref{fig:ICD_dist}\footnote{\textcolor{black}{The number documents on the y-axis, rather than the number of visits, follows the convention of \citet{mullenbach2018explainable} and is mostly used in NLP community.}}. 
The limitation of data acquisition also exacerbates the imbalance. 
The data from intensive care units such as the MIMIC-III contains severe cases with other complications. 
Patient records of the visit to general practitioners only have some general codes.

\textcolor{black}{
As an interdisciplinary study, the successful development of automated medical coding requires the collaboration between computer scientists and clinical coders~\cite{venkatesh2023automating}. 
Automated medical coding still has a long way to go with such challenging matters. 
This review illustrates the development trend in recent deep learning-based medical coding methods and proposes a unified encoder-decoder framework to shed light on future research.
We investigate how the existing methods can be categorized into the encoder-decoder framework widely adapted by many AI applications and summarize and}
review deep learning-based natural language processing techniques for automated medical coding to provide theoretical and pragmatic insights into the varieties and nuances of neural network architectures in this field.
We unify recent neural network-based methods into an encoder-decoder framework and introduce them under this unified framework. 
This review is organized as follows.
Sec.~\ref{sec:related} introduces related reviews in this field and highlights our contributions.
We formulate the unified encoder-decoder framework and corresponding building components in Sec.~\ref{sec:method}. 
Sec.~\ref{sec:benchmark} introduces widely-used benchmarks and real-world applications. 
We discuss current limitations and point out future research directions in Sec.~\ref{sec:discussion}.
Finally, we conclude our studies in Sec.~\ref{sec:conclusion}.

\section{Related Reviews and Contributions}
\label{sec:related}

There have been several systematic and narrative reviews on automated medical coding, as summarized in Table~\ref{tab:reviews}. 
One of the first reviews reported the published accuracy of discharge coding in literature~\cite{campbell2001systematic}.
Burns et al.~\cite{burns2011} conducted an updated review on the accuracy of routinely collected data following Campbell et al.~\cite{campbell2001systematic}.
Stanfill et al.~\cite{stanfill2010systematic} introduced some conventional classification methods and evaluated different types of automated coding systems.
Campbell et al.~\cite{campbell2020computer} conducted an application-oriented review of computer-assisted clinical coding.
A recent systematic review~\cite{kaur2021systematic,kaur2023ai} followed the Preferred Reporting Items for Systematic Reviews and Meta-Analyses (PRISMA) guidelines and searched publications with machine learning (ML) and natural language processing techniques. 
The authors reviewed publications from 2010 to 2020 in a narrative way. 
\textcolor{black}{Khope and Elias~\cite{khope2023strategies} conducted a similar systematic review but focused on the studies that used the MIMIC-III dataset.}
Fewer reviews covered technical matters in medical coding. 
Teng et al.~\cite{teng2022review} conducted a technical review that discusses recent advances in machine learning and natural language processing on medical coding.
Published in early 2022, it is a concurrent work focusing on feature engineering-based classifiers and deep learning methods. 
However, it does not provide a unified view that can generalize to all the nuanced varieties of deep learning architectures or cover the most recent learning algorithms for automated medical coding. 
\textcolor{black}{
Those limitations of exisiting reviews motivate us to propose a unified view of automatic medical coding models.
}

\begin{table}[htbp!]
    \centering\footnotesize
    \begin{tabular}{llll}
    \toprule
    Publications     &  Period & Category & Scope or Focus \\
    \midrule
    Campbell et al.~\cite{campbell2001systematic} & 1975 - 1998 & Systematic review & Coding accuracy \\
    Stanfill et al.~\cite{stanfill2010systematic}& 1996 - 2009 & Systematic review & Automated coding tools \\
    Burns et al.~\cite{burns2011}     & 1990 - 2010 & Systematic review & Coding accuracy\\
    Campbell et al.~\cite{campbell2020computer} & 2006 - 2017 & Narrative review & Computer-assisted clinical coding \\
    Kaur et al.~\cite{kaur2021systematic} & 2010 - 2020 & Systematic review & ML and NLP techniques\\
    Teng et al.~\cite{teng2022review}& 1990s - 2021 & Technical review & ML and NLP techniques \\
    \textcolor{black}{Khope and Elias~\cite{khope2023strategies}} & \textcolor{black}{2017-2023} & \textcolor{black}{Systematic review} & \textcolor{black}{Studies on MIMIC-III dataset}\\ 
    \textcolor{black}{Kaur et al.~\cite{kaur2023ai}} & \textcolor{black}{2010 - 2021} & \textcolor{black}{Systematic review} & \textcolor{black}{ML and NLP techniques} \\
    \hline
    Ours & \textcolor{black}{2010s - 2023} & Technical review & Unified deep learning framework\\
    \bottomrule
    \end{tabular}
    \caption{A summary of related review articles on medical coding}
    \label{tab:reviews}
\end{table}

Previous reviews introduce conventional classification systems or neural network-based methods that devise various network architectures to improve predictive performance.
However, there is no unified study on medical coding models nor an insightful analysis of the overall model architecture's submodules to solve the challenges mentioned earlier. 
Besides, recent deep learning advances beyond standard supervised learning are less discussed in existing surveys. 
The emerging deep learning paradigms include multitask learning, few-shot and zero-shot learning, contrastive learning, adversarial generative learning, and reinforcement learning.

\begin{figure}[htbp!]
\begin{center}
\includegraphics[width=0.9\linewidth]{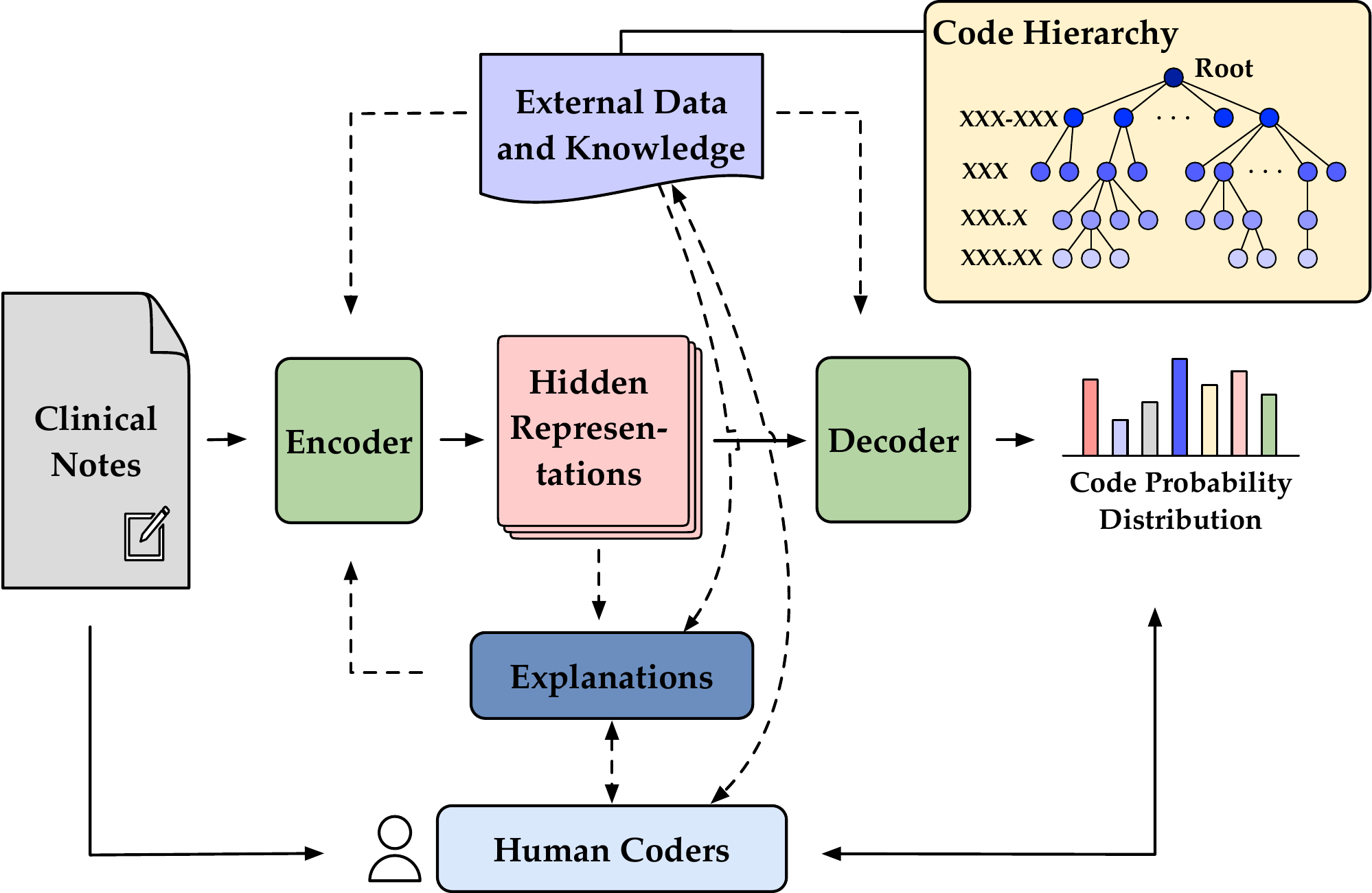}
\caption{An illustration of the unified encoder-decoder framework for automated medical coding}
\label{fig:model}
\end{center}
\end{figure}

This paper focuses on deep learning-based NLP techniques and proposes an encoder-decoder framework (Fig.~\ref{fig:model}) to unify existing advanced medical coding models. 
It discusses the effect of different building blocks to resolve the challenges of medical coding.
The categorization of building blocks is summarized in Table~\ref{tab:blocks}.
We provide a complete guideline for researchers or practitioners to develop efficient neural networks for automated medical coding and analyze the critical problems for tackling the existing challenges. 
Besides, we discuss the evaluation of medical coding and its real-world practice. 
Finally, we summarize the recent research trends and limitations and point out several vital directions for future research.  
Medical coding tasks evolve rapidly, with many deep learning-based publications emerging.
\textcolor{black}{
In this review, we curate a collelction of publications that generally come from academic databases such as PubMed, IEEE Xplore, and ACL Anthology.
}
We conduct this timely review to fill the gap by presenting a unified review and introducing recent advances in deep neural architectures for automated medical coding and emerging learning paradigms beyond supervised learning. 

\textcolor{black}{
The development of a unified encoder-decoder framework for advanced medical coding models, as discussed in this paper, serves several crucial purposes in the field of NLP applied to healthcare. 
This framework serves to unify existing models, tackle specific challenges in medical coding, and offer practical guidelines for researchers and practitioners. 
It emphasizes the importance of real-world evaluation and staying current with the rapidly evolving landscape of medical coding tasks. 
By summarizing recent trends, identifying limitations, and suggesting future research directions, the paper provides a comprehensive resource for advancing the development of efficient neural networks for automated medical coding and supports the continued growth of this critical field.
}

\begin{table}[htbp!]
    \centering\footnotesize
    \begin{tabular}{lp{4cm}p{7cm}}
    \toprule
    Categories & Functions & Representative Methods \\
    \midrule
    Encoders & Extract text features, \textcolor{black}{explainable feature learning} & CNN, RNN, graph neural networks, attention, Transformers, capsule networks \\
    Deep Connections & Build deep architecture & Stacking, residual networks, embedding injection \\ 
    Decoders & Improve code prediction & Linear layer, attention, hierarchical decoders, multitask decoders, few-shot/zero-shot decoders, \textcolor{black}{autoregressive generative decoders} \\
    Auxiliary Data & Enhance feature learning, \textcolor{black}{human-in-the-loop learning} & Code descriptions, code hierarchy, Wikipedia articles, chart data, entities and concepts, human-in-the-loop learning \\
    \bottomrule
	\end{tabular}
\caption{Categorization of building blocks under the unified framework}
\label{tab:blocks}
\end{table}

\section{A Unified Encoder-Decoder Framework}
\label{sec:method}

The recent development of automated medical coding devises novel neural networks for medical code prediction. 
For example, recurrent neural network (RNN) based methods such as the long short-term memory (LSTM) network with attention mechanism~\cite{shi2017towards} and GRU network with hierarchical attention~\cite{baumel2018multi} have been widely applied to medical code prediction from discharge summaries.  
Deep learning-based CNN models have been compared with a conventional classifier for diagnosis coding from radiology reports~\cite{karimi2017automatic}. 
In addition to conventional supervised learning, many novel learning paradigms have also been studied, for example, multitask learning~\cite{zhang2021survey} and few-shot learning~\cite{wang2020generalizing}.
This review focuses on deep learning-based NLP techniques applied to automated medical coding, unifies recent advances to introduce their advantages, and summarizes their building blocks' theoretical and pragmatic motivations.
 
We propose a unified encoder-decoder framework (Fig.~\ref{fig:model}) for automated medical coding.
\textcolor{black}{In particular, encoders refer to modules responsible for extracting relevant features from clinical text data, while decoders transform these features into medical codes.
}
The encoder modules take clinical notes as inputs and learn hidden representations, as described in Sec.~\ref{sec:encoder}.
\textcolor{black}{One important aspect of hidden representation learned by neural encoders is to produce explanations and enable trustworthy coding systems grounded by human evaluation.}
\textcolor{black}{Machine learning (ML) algorithms excel in prediction and performance but often lack transparency, emphasizing the need for explainable systems as discussed by Adadi \cite{adadi2018peeking}.
In the ML context, ``interpretability'' signifies a model's inherent understandability, while ``explainability'' refers to methods to make a model interpretable. 
Explanations are the specific insights provided by a model to aid users in understanding predictions.
A current debate revolves around whether attention mechanisms contribute to model explanation, explored by Bibal et al. \citep{bibal-etal-2022-attention}.
}
We also introduce and summarize mechanisms for deepening the architectures in Sec.~\ref{sec:building}.
The decoder modules decode the hidden representations to predict the code probability (Sec.~\ref{sec:decoder}). 
\textcolor{black}{
While the choice of encoder and decoder can be interrelated, separating these modules within the framework enhances modularity, flexibility, comparative analysis, interchangeability, and conceptual clarity. 
These benefits support a more comprehensive exploration of deep learning approaches for medical coding while accommodating the complexity of encoder-decoder interactions in various coding scenarios.
}

\textcolor{black}{In addition to decoders in the standard supervised setting, we review recent advances such as multitask decoders, few-shot/zero-shot decoders, and autoregressive generative encoders.}
During encoding and decoding, auxiliary information such as code hierarchy and textual descriptions can also be applied for enhancing representation learning and improving decoding, which is discussed in Sec.~\ref{sec:auxiliary}. 
Besides, augmented learning with external information, especially ontological knowledge, promotes explainable medical coding. 
\textcolor{black}{
Human-in-the-loop learning integrates human coders into the automated medical coding system to further enhance the encoder-decoder framework. 
They collaborate with automated algorithms to validate and improve coding accuracy. Human coders are involved in various tasks, including validating the suggestions made by automated systems, resolving ambiguous cases, ensuring compliance with coding guidelines, and refining the overall coding process. Their expertise helps refine machine learning models, contributing to ongoing system enhancement and ensuring high-quality, accurate medical coding.
For example, active learning reduces annotation costs, and human-grounded evaluation enables reliable performance evaluation. 
}
We summarize recent representative models in Table~\ref{tab:representative_models} under the proposed unified framework and review them in the following subsections.

\begin{table*}[htbp!]
\footnotesize
\begin{center}
\begin{tabular}{l | c | c| c | c }
\toprule
Models & Encoders & Deep Connections & Decoders & Auxiliary Data \\
\midrule
Attentive LSTM~\cite{shi2017towards}	&	Attentive LSTM & Stacking  & Linear Layer & NA. \\
HA-GRU~\cite{baumel2018multi} & Hierarchical GRU & Stacking &  Attention & NA. \\
LAAT~\cite{vu2020label} & BiGRU & Stacking & Attention & NA. \\
MT-RAM~\cite{sun2021multitask} & BiGRU & RAM & LAN+Multitask & NA. \\
BiCapsNetLE~\cite{bao2021medical} & BiLSTM+CapsNet & Stacking & Attention & ICD Description \\
CAML~\cite{mullenbach2018explainable}	&	CNN & Stacking & LAN & NA. \\
DR-CAML~\cite{mullenbach2018explainable}	&	CNN & Stacking & LAN & ICD Description \\
MVC-LDA~\cite{sadoughi2018medical} & Multi-view CNN & Stacking & Attention & ICD Description \\
MultiResCNN~\cite{li2020multirescnn}		&	CNN & Residual Network & LAN & NA.\\
DCAN~\cite{ji2020dilated} & Dilated CNN & Residual Network & LAN & NA. \\
HyperCore~\cite{cao2020hypercore}	&	CNN+Hyperbolic & Stacking & LAN+GCN & ICD Hierarchy \\ %
GatedCNN-NCI~\cite{ji2021medical} & Gated CNN & Embedding Injection & NCI & ICD Description \\
Fusion~\cite{luo2021fusion} & Compressed CNN & Residual Network & Attention & NA. \\
C-MemNN~\cite{prakash2017condensed}	&	Memory Networks & Stacking  & Linear Layer & NA. \\
KSI~\cite{bai2019improving} & CNN/RNN & Stacking & Linear or LAN & Wikipedia Articles \\
MCDA~\cite{wang2022novel} & CNN/RNN & Stacking & Concept-drive Attention &  Wikipedia Articles \\
MSATT-KG~\cite{xie2019ehr} & CNN+Attention & Stacking & Attention+KG & ICD Hierarchy \\
CAIC~\cite{teng2020automatic} & CNN/RNN & Stacking & Attention & ICD Description \\
GMAN~\cite{yuan2020graph} & GCN & Stacking & Mutual Attention & Patient Info.\\
JLAN~\cite{li2021jlan} & BiLSTM & Residual Network & Self-attention+LAN & ICD Description \\
DACNM~\cite{cao2020clinical} & Dilated CNN & Stacking & N-gram+Linear & ICD Description \\
BERT-XML~\cite{zhang2020bert} & BERT & Stacking & LAN & ICD Description \\
ISD~\cite{zhou2021automatic} & CNN & Stacking & Attention & NA. \\
CMGE~\cite{wu2021counterfactual} & Graph encoder & Stacking & Multitask Decoder & NA. \\
RAC~\cite{kim2021read} & CNN & Stacking & Attention & Data Augmentation \\
ICDBigBird~\cite{michalopoulos2022icdbigbird} & BigBird & Stacking &  Label Attention & NA. \\ 
HieNet~\cite{wang2022hienet} & CNN & Stacking & Progressive Mechanism & ICD Hierarchy \\
MD-BERT~\cite{zhang2022hierarchical} & Hierarchical BERT & Stacking & Label Attention & ICD Description \\
MSMN~\cite{yuan2022code} & LSTM & Stacking & Multi-synonyms Attention & UMLS \\
\bottomrule
\end{tabular}
\end{center}
\caption{A summary of representative models under the unified encoder-decoder framework}
\label{tab:representative_models}
\end{table*}

\subsection{Encoder Modules}
\label{sec:encoder}
Deep learning-based models use word embedding techniques and develop complex neural network architectures to learn rich text features for automatic medical code assignment.
After some text preprocessing techniques, a clinical note with $n$ words is denoted as $\{x_0, \dots, x_n\}$. 
Its word embedding matrix, for example, built by word2vec~\cite{mikolov2013distributed} or GloVe~\cite{pennington2014glove}, is denoted as $\mathbf{X}=[ \mathbf{w}_1, \dots, \mathbf{w}_n]^T\in \mathbb{R}^{n\times d_e}$, where $d_e$ is the dimension of word vectors. 
The encoder modules of various neural architectures further process embeddings to learn rich hidden representations. 
\textcolor{black}{
In the context of deep learning for medical coding, CNN-based approaches focus on extracting local features and patterns from clinical documents, while RNN-based methods are employed to capture sequential dependencies and contextual information within the text. 
Recent publications also leverage advanced language models like BERT, which provide contextualized word embeddings, enhancing the understanding of medical narratives by considering the broader context of each word in the document. 
These diverse techniques contribute to more accurate and comprehensive automated medical coding systems.
}
This section introduces various neural encoder modules that have been developed in recent years.

\subsubsection{Recurrent Neural Encoders}
\label{sec:RNN}

Recurrent neural networks model the temporal sequences via their internal states and capture sequential dependencies. 
Thus, they have been widely applied to textual sequence modeling and clinical note encoding. 
Generally, the recurrent neural encoder (in Fig.~\ref{fig:rnn_encoder}) outputs a hidden representation $\mathbf{H}^l \in \mathbb{R}^{n\times d_h}$ of the $l$-th layer:
 \begin{equation}
 	\mathbf{H}^l = \operatorname{RNN}(\mathbf{X}),
 \end{equation}
where $n$ is the number of words and $d_h$ is the dimension of the hidden representation.
However, the vanilla RNN-based model suffers from the vanishing gradient issue~\cite{hochreiter1998vanishing}. 
Shi et al.~\cite{shi2017towards}, one of the first works on applying RNNs for medical coding, developed an Attentive LSTM network.
This model encodes clinical descriptions and long titles of ICD codes jointly with hierarchical text representations and uses an attention mechanism for matching important diagnosis snippets. 
Catling et al.~\cite{catling2018towards} compared the TF-IDF (term frequency-inverse document frequency) feature with the word embedding features learned with the simplified gated recurrent unit (GRU).
Mullenbach et al.~\cite{mullenbach2018explainable} used a GRU with bi-direction as a baseline system for medical coding, where the last hidden representations are used for classification. 
From their pilot experiments, GRU shows more robust predictive performance than the LSTM network-based coding model.  
\textcolor{black}{
\citet{blanco2020extreme} studied capabilities of various RNN models such as GRU and ELMo (Embeddings from Language Model).
}
Other follow-up works such as HA-GRU~\cite{baumel2018multi} and HLAN~\cite{dong2020explainable} further improved the vanilla BiGRU with hierarchical attention, including two levels on sentence and document representations. 
Hierarchical attention can help mitigate the difficulty of encoding long text sequences. 
Sec.~\ref{sec:hierachical_encoder} introduces more details of hierarchical encoders.

\subsubsection{Convolutional Neural Encoders}
\label{sec:CNN}

The success of convolutional neural networks in computer vision inspires researchers to use convolutional architecture for medical coding.
The TextCNN model~\cite{kim2014convolutional} acts as a simple but essential baseline. 
The convolutional layer extracts local features from pretrained or randomly initialized word vectors.
Fig.~\ref{fig:cnn_encoder} illustrates the CNN-based text encoder.
The representation with max-pooling is then used for medical code classification. 
Karimi et al.~\cite{karimi2017automatic} compared standard CNN architecture with conventional classifiers such as decision trees and support vector machines~\cite{farkas2008automatic,suominen2008machine} on both in-domain and out-of-domain data and showed that CNN architectures with optimal parameter settings gain comparable results with conventional methods on sparse and skewed data. 
CAML~\cite{mullenbach2018explainable} combines multiple-filter CNN-based text encoders and an attention decoder (introduced in Sec.~\ref{sec:decoder}).  
DCAN~\cite{ji2020dilated} develops dilated convolution layers, which apply convolutions with dilated filters to increase the receptive field.  
 Given a sequence of one-dimensional elements $\mathbf{x}\in \mathbb{R}^n$ and a convolutional filter $f:\{0, \ldots, k-1\} \rightarrow \mathbb{R}$, 
 the hidden representation in the $l$-th layer of stacked dilated convolution layers is calculated as $\mathbf{H}^l_{ij} = \sum_{j=0}^{n-1} f(j) \cdot \mathbf{x}_{s-d_l \cdot j}$,
 \begin{equation}
 \mathbf{H}^l_{ij} = (\mathbf{w}_i*_{d_l} f) (\mathbf{w}_{ij}) = \sum_{j=0}^{n-1} f(j) \cdot \mathbf{x}_{s-d_l \cdot j},
 \end{equation}
 where $d_l$ is the dilation size of the spacing between kernel elements in the $l$-th layer, $s$ is the element of input sequence, and $s-d_l\cdot i$ refers to past time steps. 
When stacking a deeper architecture, the dilation size is exponentially increased to expand the receptive field.
Other models also use the CNN-based text encoder. 
For example, MultiResCNN~\cite{li2020multirescnn} concatenates the features of multi-filter convolutions. 
Similarly, MVC-LDA~\cite{sadoughi2018medical} introduces multi-view CNN by applying max-pooling over different channels with different convolutional filters.  
Ji et al.~\cite{ji2021medical} developed a Gated CNN encoder that uses an LSTM-style gating mechanism to control the information flow.
The Fusion model~\cite{luo2021fusion} deploys a Compressed CNN module that applies attention-based soft-pooling over word convolution features, reducing the number of word representations. 
Critical entities can help to recognize the correct medical code.
ECNN~\cite{chen2020towards} enhances the CNN model with entities extracted from the input text.
Inspired by the squeeze-and-excitation network~\cite{hu2018squeeze}, EffectiveCAN~\cite{liu2021effective} stacks multiple residual squeeze-and-excitation blocks with convolutional operations. 

\subsubsection{\textcolor{black}{Neural Attention and Transformer Encoders}}
\label{sec:attention}

\begin{figure}[htbp!]
\begin{center}
\begin{subfigure}[b]{0.28\textwidth}
	\includegraphics[width=\linewidth]{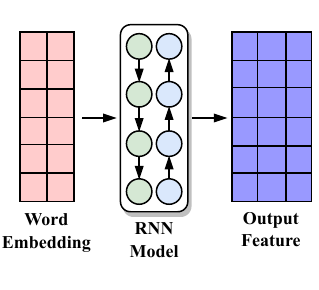}
	\caption{RNN encoder}
	\label{fig:rnn_encoder}
\end{subfigure}
\qquad
\begin{subfigure}[b]{0.28\textwidth}
	\includegraphics[width=\linewidth]{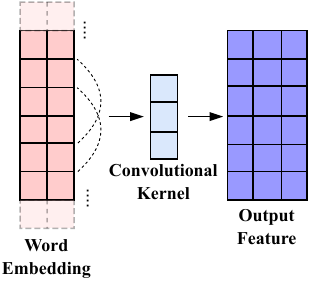}
	\caption{CNN encoder}
	\label{fig:cnn_encoder}
\end{subfigure}
\qquad
\begin{subfigure}[b]{0.28\textwidth}
	\includegraphics[width=\linewidth]{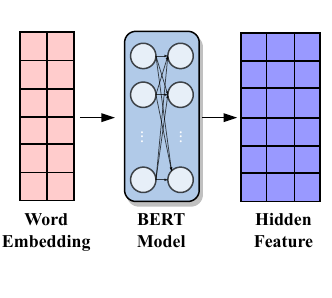}
	\caption{BERT encoder}
	\label{fig:bert_encoder}
\end{subfigure}
\caption{Illustrations of representative neural text encoders. (a) The RNN encoder captures sequential dependency. (b) The CNN encoder extracts local features. (c) The BERT encoder encodes contextualized information.}
\label{fig:}
\end{center}
\end{figure}

The neural attention mechanism computes a weighted sum of vector values of hidden representations dependent on the query vectors.
Compared to RNN and CNN, self-attention has been widely adopted for transfer learning, i.e., as building blocks for large pre-trained language models. 
This allows leveraging the linguistic associations from massive corpora for subsequent tasks.
The superior performance gained by BERT attracts researchers of medical coding to apply BERT-based text encoders, as shown in Fig.~\ref{fig:bert_encoder}.
\textcolor{black}{
However, due to the complexity of the self-attention mechanism, only a few works use pure attention-based encoders to model the clinical notes especially discharge summaries. 
Since 2021, more researchers have proposed to use transformer-based models.
}
TransICD~\cite{biswas2021transicd} applies transformer text encoder and structured self-attention to learn representations.
\textcolor{black}{
Coutinho et al.~\cite{coutinho2022transformer} used Transformers for ICD-10 coding from Portuguese text.
}
Some attempts explore the possibility of BERT encoders. 
\textcolor{black}{
\citet{roitero2021dilbert} built a BERT model via domain-specific pretraining and fine-tuning.
}
\textcolor{black}{BERT encoders are limited to encoder the maximum sequence length of 512.
When dealing with long documents, they do not achieve superior performance compared with CNN or RNN-based encoders, potentially due to the limitation of BERT to encode long documents and keywords according to Gao et al.~\cite{gao2021limitations}. 
Thus, BERT encoders are usually used to encode the long clinical notes in a hierarchical manner, which will be introduced in Sec.~\ref{sec:hierachical_encoder}.  
}

More recent studies attempt to study the performance of efficient transformer-based methods. 
For example, Feucht et al.~\cite{feucht2021description} found that Longformer achieves better results than BERT.
Yogarajan et al.~\cite{yogarajan2021improving} applied concatenated representations from contextualized language models and used Longformer~\cite{beltagy2020longformer} and Transformer-XL~\cite{dai2019transformer} to processed longer sequences. 
\textcolor{black}{
Yang et al.~\cite{yang2022knowledge} adopted longformer with domain-specific knowledge enhancement.
} \textcolor{black}{Hou et al. \cite{hou2024modelling} integrated the long-distance dependency features captured through Clinical-Longformer with code synonyms, code hierarchy, and code co-occurrence knowledge to improve long-tail classification. Gomes et al. \cite{gomes2024accurate} compared the ability of the chunk encoder and longformer encoder for lengthy text modeling. }
Michalopoulos et al.~\cite{michalopoulos2022icdbigbird} applied BigBird~\cite{zaheer2020big} designed for long sequence encoding to encode discharge summaries. 
Niu et al. \cite{niu2023retrieve} used the FLASH \cite{hua2022transformer}, a variant of Transformer, as a feature extractor to extract meaningful semantic features from long clinical notes.
Liu et al. \cite{liu2022hierarchical} pre-trained the new language model ClinicalplusXLNet based on fine-tuning the pre-trained Transformer model. The authors conducted continuous pre-training using clinical corpus from MIMIC-III using XLNet-Base. Subsequently, Duan et al. \cite{duan2023mhlat} employed ClinicalplusXLNet as the encoder, encoding the segmented clinical text to obtain semantic features.
\textcolor{black}{Xie et al. \cite{xie2024knowledge} developed a knowledge-based dynamic prompt learning algorithm for coding prediction. The method utilizes various masked language models and dynamically generates prompts based on personal medical information and medical knowledge graphs to provide valuable information representation for the model training.}

Previous methods rely on existing pretrained language models to obtain contextualized embeddings.
Zhang et al.~\cite{zhang2020bert} proposed BERT-XML that combines BERT encoders with multi-label attention. 
Rather than fine-tuning the pretrained BERT encoder, the authors trained the self-supervised BERT-XML encoder from scratch on clinical notes to solve the out-of-vocabulary issue. 
Moreover, they pretrained the BERT-XML model with a sequence length of 1024 for long sequences. 

\subsubsection{Graph Encoders}
\begin{figure}[htbp!]
\begin{center}
\begin{subfigure}[b]{0.4\textwidth}
	\includegraphics[width=\linewidth]{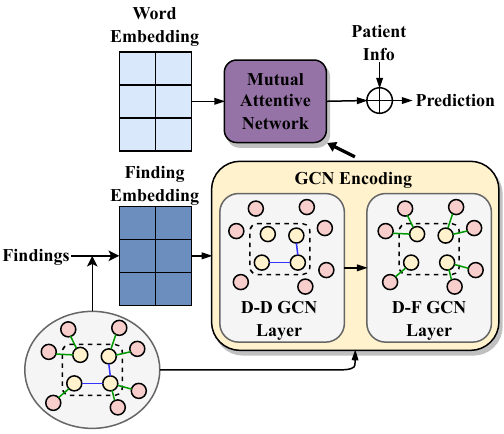}
	\caption{GMAN~\cite{yuan2020graph}}
	\label{fig:graph_GMAN}
\end{subfigure}
\qquad
\begin{subfigure}[b]{0.4\textwidth}
	\includegraphics[height=4.5cm]{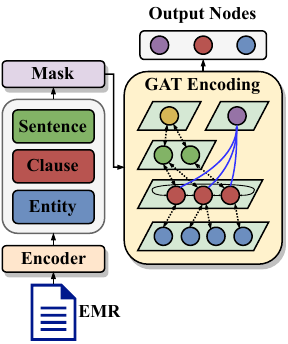}
	\caption{CMGE~\cite{wu2021counterfactual}}
	\label{fig:graph_CMGE}
\end{subfigure}
\caption{Graph encoders on the constructed medical graph. a) GMAN applies the GCN layers to encode disease hierarchy and disease-finding causal relations. b) CMGE uses the GAT as the graph encoder on a multi-granularity graph.}
\label{fig:graph_encoders}
\end{center}
\end{figure}

Many natural language processing tasks construct text graphs and adopt graph neural networks as text encoders to learn textual features~\cite{wu2023graph}.
Several works in medical coding also use graph-based encoders to capture the structural information during the diagnosis process.
Yuan et al.~\cite{yuan2020graph} built a medical graph that consists of diseases and findings and deployed the Graph Convolutional Network (GCN)~\cite{kipf2017semi} to learn graph representations. 
The authors considered disease-disease (D-D) and disease-finding (D-F) graphs as shown in Fig.~\ref{fig:graph_GMAN} during the encoding of disease hierarchy and causal relations. \textcolor{black}{Similarly, Lu et al. \cite{lu2023combining} transformed text features extracted by pre-trained BERT into node representations in heterogeneous graphs and utilized GCN for message passing.} 
CMGE~\cite{wu2021counterfactual}, a multi-granularity graph-based method, builds a hierarchical graph that contains four types of nodes: general nodes (patients' age and gender), sentence nodes, clause nodes, and entity nodes, as illustrated in Fig.~\ref{fig:graph_CMGE}.
It uses the Graph Attention Network (GAT)~\cite{velivckovic2018graph} for information aggregation.
The multi-granularity graph reasoning enables supporting fact extraction from the clinical notes and explainable diagnosis prediction. \textcolor{black}{Luo et al. \cite{luo2024corelation} constructed a code relation graph to capture the complex interaction relationships between ICD codes and improve code allocation accuracy.}

\subsubsection{Hierarchical Encoders}
\label{sec:hierachical_encoder}
\begin{figure}[htbp!]
\begin{center}
\begin{subfigure}[b]{0.5\textwidth}
	\centering
	\includegraphics[height=4cm]{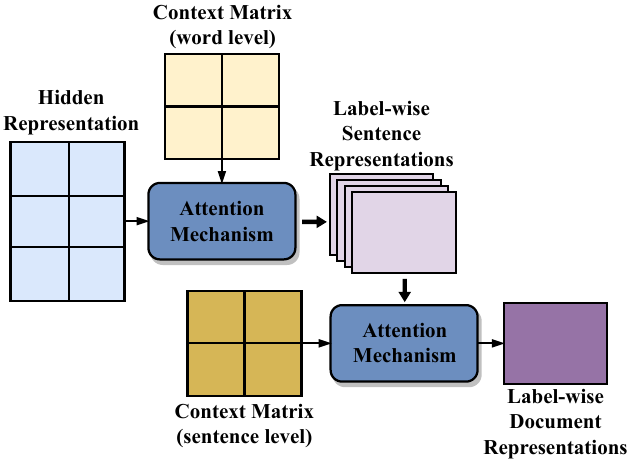}
	\caption{Hierarchical encoding~\cite{dong2020explainable}}
	\label{fig:hier-encoding}
\end{subfigure}
\qquad
\begin{subfigure}[b]{0.3\textwidth}
	\includegraphics[height=4cm]{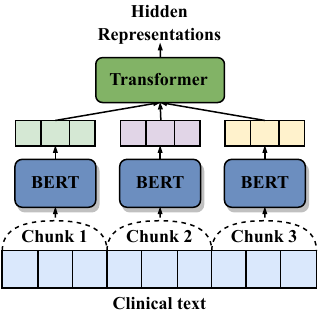}
	\caption{Hierarchical BERT~\cite{ji2021does}}
	\label{fig:hier-bert}
\end{subfigure}
\caption{Illustrations of the hierarchical encoder and decoder. (a) The hierarchical encoder learns hierarchical representations of words, sentences, and documents. (b) The hierarchical BERT encoder processes text chunk by chunk and aggregates the embeddings via an additional Transformer.}
\label{fig:hierarchical}
\end{center}
\end{figure}

Several methods adopt hierarchical text encoders, as a ``meta''-encoder with the above encoding modules, to take the hierarchical structure of documents into encoding and potentially solve the difficulty in encoding lengthy clinical documents that encode hierarchical elements of the long documents such as characters, words, sentences, and chunks.
Shi et al.~\cite{shi2017towards} built a hierarchical encoder with character representation, word representation, and sentence representation.
Dong et al.~\cite{dong2020explainable} adapted hierarchical attention networks with label-wise word-level and sentence-level representations for an improved attention-based explanation for each code (in Fig.~\ref{fig:hier-encoding}).
To make BERT-based text encoders compatible with long clinical notes, Ji et al.~\cite{ji2021does} developed BERT-hier that divides long notes into chunks and uses another Transformer network to encode the embeddings of different chunks (in Fig.~\ref{fig:hier-bert}).
Although the hierarchical BERT-based encoder improves the performance, it is still not as good as advanced CNN or RNN-based models.
\textcolor{black}{
Pascual et al.~\cite{pascual2021towards} conducted a similar study.
A recent work called Medical Document BERT (MD-BERT)~\cite{zhang2022hierarchical} proposes a more advanced hierarchical encoding method by considering token-level, sentence-level, and document-level representation learning and attaching the classification layer to any levels of interest according to specific tasks. 
This model achieves better performance than previous attempts on utilizing transformers-based text encoders. 
Other recent findings also show that BERT-based encoders can achieve improved performance with better configuration and training when handling long texts.
For example, Dai et al.~\cite{dai2022revisiting} showed that the document splitting strategy for text encoders is important. 
Afkanpour et al.~\cite{afkanpour2022bert} found that the utilization of token-level representation and longer text sequence can improve performance.
In addition to discharge summaries, text metadata such as time and note type is also used in the hierarchical transformer model to improve temporal document sequence encoding~\cite{ng2023modelling}.
}

\subsubsection{Summary}
Neural encoders play an important role in learning rich representations from clinical notes.
Early research on deep learning-based medical coding explored recurrent and convolutional neural networks and achieved improved performance than feature engineering-based methods. 
Many follow-up works further improve CNNs and RNNs to enhance their capacity to capture long context. 
Self-attention-based Transformer networks suffer from quadradic complexity. 
However, recent studies on hierarchical encoders and efficient transformers are getting a better performance for medical coding.
Graph neural networks that can capture structural information represented in heterogeneous text graphs are emerging. 
The inductive bias in different neural encoders is the key consideration for the choice of encoders.
However, there is no clear evidence on which neural encoder is optimal. 
One recommendation is to choose the neural encoder based on the data and the need for coding practice. 
\textcolor{black}{
Neural architectures offer benefits such as automated feature learning, representation hierarchies, and performance improvement in medical coding tasks. 
However, their black-box nature, model complexity, lack of intuitive representations, and the need for additional explainability techniques pose significant challenges in achieving interpretability and explainability. 
Future research requires to develop methods to address these limitations and make neural architectures more transparent and understandable for medical coding applications.
}

\subsection{Building Deep Architectures}
\label{sec:building}
Most existing neural network-based medical coding models have deep architectures.
The most straightforward approach uses stacking to build deep neural architectures, such as stacking multiple recurrent layers and hierarchical components of different levels of elements as in multi-layer perceptrons.  
Also, different neural blocks can be stacked into deep networks, for example, the recalibrated aggregation module~\cite{sun2021multitask} with multiple convolutional layers is built upon a bidirectional GRU network, the MSATT-KG~\cite{xie2019ehr} stacks densely connected convolutional layers and multi-scale feature attention, and the BiCapsNetLE~\cite{bao2021medical} deploys a capsule neural network upon the BiLSTM layer to extract features further.

When encoding long clinical notes with very deep architectures, features learned by higher layers tend to capture abstract features but sometimes miss some vital information. 
Ji et al.~\cite{ji2021medical} proposed to use embedding injection to mitigate the information loss with the increase of neural layers. 
The embedding injection concatenates the original word embeddings into each intermediate layer of the backbone network as: 
\begin{equation}
 	\mathbf{J}^{l}=\text{concat}\left[\mathbf{X},~\mathbf{H}^{l}\right],
 \end{equation}
where $\mathbf{J}^{l} \in \mathbb{R}^{n \times\left(d_{e}+d_{h}\right)}$ are the features with original embeddings injected.

\begin{figure}[htbp!]
\begin{center}
\begin{subfigure}[b]{0.4\textwidth}
	\includegraphics[width=\linewidth]{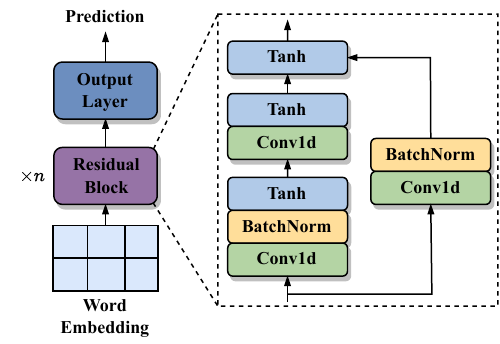}
	\caption{Residual networks}
	\label{fig:residual_networks}
\end{subfigure}
\qquad
\begin{subfigure}[b]{0.4\textwidth}
	\includegraphics[width=\linewidth]{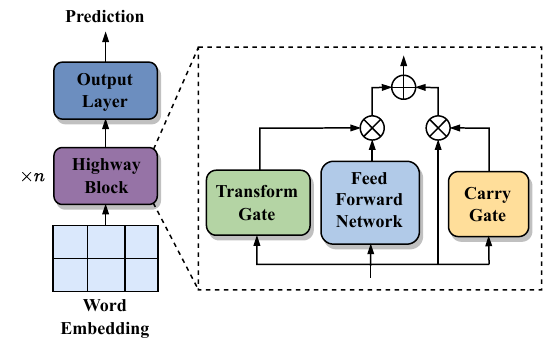}
	\caption{Highway networks}
	\label{fig:highway_networks}
\end{subfigure}
\caption{Illustrations of residual networks and highway networks for building deep architectures}
\label{fig:deep_architecture}
\end{center}
\end{figure}

The most widely used approach to building deep networks for automated medical coding is to use residual connections, as shown in Fig.~\ref{fig:residual_networks}.
Deep residual learning introduces the skip connection to avoid the effect of the vanishing gradient. 
It enables the building of very deep neural network architectures. 
Given the input encoding vector $\mathbf{x}$, the output of residual connection is denoted as $o=\sigma(\mathbf{x}+\mathcal{G}(\mathbf{x}))$, where $\mathcal{G}$ represents neural layers and $\sigma$ is a non-linear activation function.
Several medical coding models use residual networks between stacked layers, which are denoted as: 
 \begin{equation}
 \mathbf{H}^{l+1} = \sigma(\mathbf{H}^l+\mathcal{G}(\mathbf{H}^l)).
 \end{equation}
MultiResCNN~\cite{li2020multirescnn} is the first to combine residual learning with the concatenation of multiple channels with different convolutional filters.
Other follow-up works such as DCAN~\cite{ji2020dilated} and Fusion~\cite{luo2021fusion} also use the residual neural network.
We also illustrate the highway networks for building deep architectures in Fig.~\ref{fig:highway_networks}, although no existing medical coding models adopt the highway mechanism. 
Highway networks use the gating mechanism (i.e., the transform gate and the carry gate) to control the amount of input information and avoid attenuation when stacking very deep layers.
The highway networks can be an alternative to building deep medical coding models. 

\subsection{Decoder Modules}
\label{sec:decoder}
After the encoder modules have extracted hidden representations of clinical notes, the decoder modules map the learned representations into medical codes as the final classification results via a decoding process.
The hierarchical and large-scale characteristics of medical codes have promoted the design of various decoder modules.
\textcolor{black}{
This section introduces four main types of decoder modules, including the fully connected layer-based decoder that offers simplicity and efficiency (Sec.~\ref{sec:decoder_FCN}), neural attention decoders that enhance code prediction by focusing on relevant information (Sec.~\ref{sec:decoder_attention}), hierarchical decoders that leverage the hierarchical structure of medical codes for more structured predictions (Sec.~\ref{sec:decoder_hier}), multitask decoders that handle multiple coding systems simultaneously (Sec.~\ref{sec:decoder_multitask}), and few-shot decoders aim to solve the few-shot learning problem and make accurate predictions with minimal examples, making them valuable in challenging coding situations (Sec.~\ref{sec:decoder_fewshot}). 
}

In addition to these decoders, pipeline-based methods further deploy some post-hoc modules to boost performance. 
For example, Tsai et al.~\cite{tsai2021modeling} proposed a two-stage method, i.e., a candidate generation stage to generate candidate sets of ICD codes and a candidate reranking stage that leverages the label correlation to rerank the generated code sets. 
Some non-parametric post-processing methods can also be applied for adjusting the decoders of medical coding models; for example, the Classification with Alternating Normalization method~\cite{jia2021doubt} that redistributes the prediction probability. 

\subsubsection{Fully Connected Layer}
\label{sec:decoder_FCN}
The most straightforward decoder module is a linear fully-connected layer, widely used in many classification tasks. 
The prediction logits $\hat{{y}}\in \mathbb{R}^m$ between 0 and 1 are produced by the $\operatorname{Sigmoid}$ activation function with a pooling operation over the linearly projected matrix, calculated as:
 \begin{equation}
 \hat{{y}}=\operatorname{Sigmoid}(\operatorname{Pooling}(\mathbf{H} \mathbf{W}^\mathrm{T}),
 \end{equation}
where $\mathbf{W}\in \mathbb{R}^{m\times d_h}$ are the linear weights for $m$ medical codes.
Medical coding models that use a linear layer as a decoder include Attentive LSTM~\cite{shi2017towards} and C-MemNN~\cite{prakash2017condensed}.

\subsubsection{Neural Attention Decoders}
\label{sec:decoder_attention}
\begin{figure}[htbp!]
\begin{center}
\begin{subfigure}[b]{0.4\textwidth}
	\includegraphics[height=4cm]{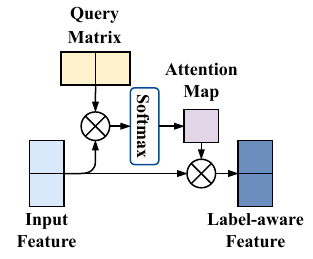}
	\caption{Label attention}
	\label{fig:LAN}
\end{subfigure}
\qquad
\begin{subfigure}[b]{0.4\textwidth}
	\includegraphics[height=4cm]{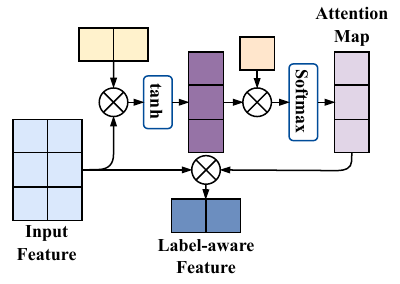}
	\caption{Structured self-attention}
	\label{fig:ssa}
\end{subfigure}
\caption{Illustrations of attention encoders. (a) Label attention learns label-aware representations for decoding. (b) Structured self-attention also learns label-specific representations.}
\label{fig:attention_decoder}
\end{center}
\end{figure}

The neural attention mechanism has also been applied to decoding in addition to its usage for encoding clinical notes introduced in Sec.~\ref{sec:attention}.
One useful attention mechanism for decoding is the so-called Label-wise Attention Network (LAN) which prioritizes important information in the hidden representation relevant to medical codes. 
The LAN-based decoder, as illustrated in Fig.~\ref{fig:LAN}, uses the dot product attention to calculate the attention score $\mathbf{A}\in \mathbb{R}^{n\times m}$ as:
 \begin{equation}
 \mathbf{A}=\operatorname{Softmax}(\mathbf{H} \mathbf{U}),
 \end{equation}
where $\mathbf{U}\in \mathbb{R}^{{h_L}\times m}$ is the query matrix of the label attention layer for $m$ medical codes, and $h_L$ is the dimension of the query. 
By multiplying attention $\mathbf{A}$ with the hidden representation, i.e., $\mathbf{A}^{\operatorname{T}} \mathbf{H}$, the output of the attention layer is obtained for medical code prediction.
CAML~\cite{mullenbach2018explainable} is the first to apply LAN by using the attention matrix to capture the importance of ICD code and hidden word representation pair. 
DCAN~\cite{ji2020dilated} and MultiResCNN~\cite{li2020multirescnn} also use the LAN decoder as a building block of their models. 
Fusion~\cite{luo2021fusion} deploys similar code-wise attention after feature aggregation and RAC~\cite{kim2021read} implements the code-title guided attention module.
The LAN-based decoder preserves sequential information captured by the text encoder and enables label awareness to benefit medical code classification. 
JLAN~\cite{li2021jlan} proposes a dual attention mechanism that combines self-attention and label attention.
LAAT~\cite{vu2020label} applies the structured self-attention~\cite{lin2017structured} (in Fig.~\ref{fig:ssa}) that projected the hidden representation via a linear transformation and non-linear activation as:
 \begin{equation}
 	\mathbf{H}^\prime=\tanh (\mathbf{W}_s \mathbf{H}),
 \end{equation}
where $\mathbf{W}_s \in \mathbb{R}^{h\times d_h}$ is a weight matrix, and $h$ is the number of hops of the structured self-attention.
In practice, LAAT sets the number of attention hops to the number of labels.
Similarly, TransICD~\cite{biswas2021transicd} uses the structured self-attention mechanism to achieve code-specific decoding for automated medical code prediction.
Attention-based decoders enhance medical code prediction by modeling code information. 
However, due to the lack of training data, code interaction cannot be effectively learned, especially for those rare codes.
Zhou et al.~\cite{zhou2021automatic} proposed an interactive shared representation network to enhance the interaction among code-relevant information via multi-layer transformer decoders. 
To capture code co-occurrence, the authors further implemented two additional tasks, i.e., missing code completion and wrong code removal.
\textcolor{black}{
Wu et al. \cite{wu2022jan} designed a joint attention decoder that utilizes document-based attention to extract text information and label-based attention to emphasize the semantic connection between label semantics and document content.  To balance the contribution of document-based attention and label-based attention to label feature representation, the authors employed layer and gate mechanisms to achieve adaptive fusion. 
Due to the existing label attention mechanism to identify critical segments in the entire text at once, it may ignore some crucial local information scattered in paragraphs. Kim et al. \cite{kim2022automatic} designed a new neural decoder composed of two label attention layers by integrating traditional and partition-based label attention mechanisms to obtain global and local potential feature representations. Partition-based label attention divides the text representation obtained from the encoder and generates label-specific features for each segment. Then, the weighted summation of features is performed to obtain a combined label-specific feature matrix.
\textcolor{black}{Inspired by the coding process of clinical coders (selecting general categories first and then specific subcategories), Nguyen et al. \cite{nguyen2023twostage} designed a two-stage decoding process that utilizes attention mechanisms first to predict the parent code and then the child code based on previous predictions.}
}

\subsubsection{Hierarchical Decoders}
\label{sec:decoder_hier}
\begin{figure}[htbp!]
\begin{center}
	\includegraphics[height=4cm]{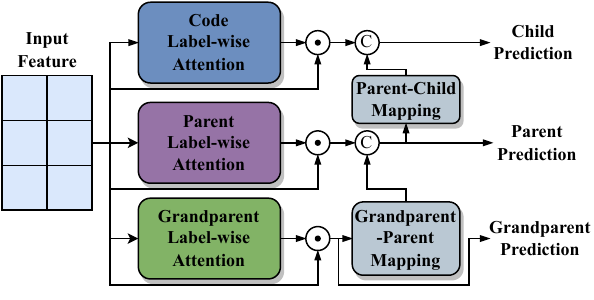}
\caption{Illustrations of the hierarchical decoder that considers the hierarchical structure of medical ontology during code prediction \cite{falis2019ontological}. Circles with dots represent dot production and circles with ``C'' denotes concatenation.}
\label{fig:hierarchical_decoder}
\end{center}
\end{figure}

Hierarchical models that use the hierarchical code structure have been studied to improve automatic coding a long time ago~\cite{de1998hierarchical}. 
Building hierarchical decoders is still a promising research direction in the recent advances in deep learning-based methods. 
JointLAAT~\cite{vu2020label} proposes a hierarchical joint learning method that produces the code prediction level by level according to the ICD hierarchy. 
Firstly, the model predicts the normalized ICD codes with the first three characters.
Then, the first level's predictions are projected back to a vector and concatenated with the label-specific representation of the second level in the ICD hierarchy for the final prediction. 
An earlier work by Falis et al.~\cite{falis2019ontological} uses three hierarchical decoding layers for ICD codes, as shown in Fig.~\ref{fig:hierarchical_decoder}.
The hierarchical decoding-based JointLAAT outperforms the vanilla LAAT slightly in some evaluation metrics. 
There is still room for improvement by making use of the hierarchical nature of the medical coding system.
\textcolor{black}{
Unlike those hierarchical decoding methods via joint learning, RPGNet~\cite{wang2020coding} formulates the medical coding task as a path generation problem and proposes a coarse-to-fine ICD path generation model based on adversarial reinforcement learning. 
It leverages a path generator to generate paths and a path discriminator to distinguish the generated paths from positive paths.
}
\textcolor{black}{Liu et al.~\cite{liu2022automated} proposed hierarchical label-wise attention in response to the hierarchical encoding at token and chunk levels.
}

\subsubsection{Multitask Decoders}
\label{sec:decoder_multitask}
\begin{figure}[htbp!]
\begin{center}
\begin{subfigure}[b]{0.45\textwidth}
    \centering
	\includegraphics[height=4cm]{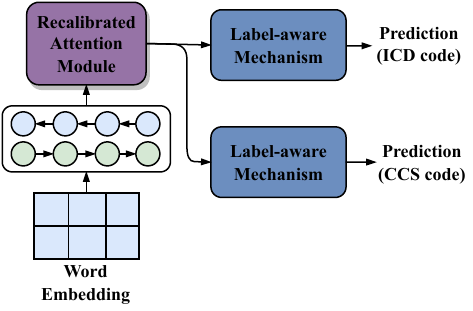}
	\caption{MT-RAM \cite{sun2021multitask}}
	\label{fig:ICD_CCS}
\end{subfigure}
\qquad
\begin{subfigure}[b]{0.45\textwidth}
    \centering
	\includegraphics[height=4cm]{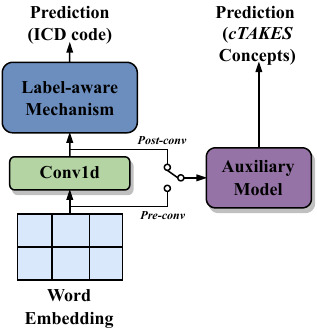}
	\caption{Wiegreffe et al.~\cite{wiegreffe2019clinical}}
	\label{fig:cTAKES}
\end{subfigure}
\qquad
\begin{subfigure}[b]{0.45\textwidth}
    \centering
 	\includegraphics[width=\linewidth]{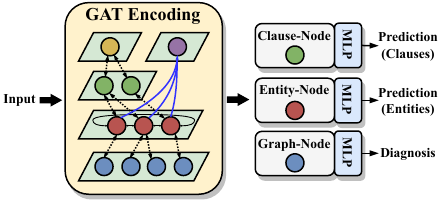}
	\caption{CMGE~\cite{wu2021counterfactual}}
	\label{fig:CMGE}
\end{subfigure}
\caption{Illustrations of multitask decoders. (a) MT-RAM adopts two-branch joint multitask training; (b) Wiegreffe et al. used distantly surprised cTAKES output prediction as a task head; (c) CMGE adopts three classification tasks for graphs, sub-sentences, and entities.}
\label{fig:multitask_decoders}
\end{center}
\end{figure}

Multitask decoders predict medical codes with multiple task branches powered by multitask learning~\cite{caruana1997multitask}. 
Tsai et al.~\cite{tsai2019leveraging} took low-level code and high-level category as two task branches in their multitask learning framework.
Medical coding models aforementioned in this section build decoders for a single coding system. 
However, several different systems have been used for different purposes. 
To enable decoding of multiple coding systems and utilize the joint learning of similar tasks, MT-RAM~\cite{sun2021multitask} deploys a multitask decoding scheme that includes two branches with label-wise attention for ICD and CCS code prediction as shown in Fig.~\ref{fig:ICD_CCS}.
As a following-up work, MARN~\cite{sun-marn} improves the multitask decoders with the focal loss to balance the learning of codes with imbalanced code frequencies. 
More publications introduce other auxiliary tasks to train joint learning models.
Wiegreffe et al.~\cite{wiegreffe2019clinical} predicted the outputs of the Apache clinical Text Analysis Knowledge Extraction System (cTAKES)\footnote{Available at \url{https://ctakes.apache.org/}} together with ICD codes as illustrated in Fig.~\ref{fig:cTAKES}. 
CMGE~\cite{wu2021counterfactual} in Fig.~\ref{fig:CMGE} considers graph classification, sub-sentence classification, and entity classification.
Rios et al.~\cite{rios2021assigning} jointly trained a multitask learning model with losses for topography and histology codes and the hierarchical loss with a hierarchical regularization.

\subsubsection{Few-shot/Zero-shot Decoders}
\label{sec:decoder_fewshot}
\begin{figure}[htbp!]
\begin{center}
\begin{subfigure}[b]{0.4\textwidth}
	\includegraphics[height=4cm]{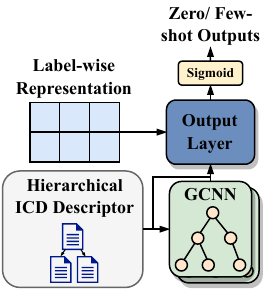}
	\caption{ZAGCNN~\cite{rios2018few}}
	\label{fig:ZAGCNN}
\end{subfigure}
\qquad
\begin{subfigure}[b]{0.4\textwidth}
	\includegraphics[width=\linewidth]{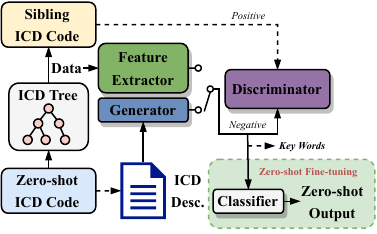}
	\caption{AGM-HT~\cite{song2020generalized}}
	\label{fig:AGM-HT}
\end{subfigure}
\caption{Illustrations of few-shot and zero-shot medical coding. (a) ZAGCNN considers few-shot/zero-shot medical coding as a retrieval problem. (b) AGM-HT utilizes adversarial generative training.}
\label{fig:few-shot}
\end{center}
\end{figure}

The automatic medical coding task has a large label space, with some frequently appearing and many labels never shown in the dataset.  
Few-shot models aim to predict codes that only appear a few times in the training data, and zero-shot models aim to predict codes that never appear in the training data.  
Rios and Kavuluru \cite{rios2018few} are among the first work of medical coding in the few-shot and zero-shot settings. 
They defined the few-shot and zero-shot coding problem as a retrieval task (in Fig.~\ref{fig:ZAGCNN}) in which the model calculates the code probability as the semantic matching between the representations of clinical documents and label vectors of target codes, denoted as:
\begin{equation}
	\tilde{y_i} = \operatorname{Sigmoid}(\mathbf{e}_i^\top \mathbf{v}_i),
\end{equation} 
where $\mathbf{e}_i$ is the label-specific document vector and $\mathbf{v}_i$ is the label vector for $i$-th label.
The proposed model ZAGCNN uses CNN layers to extract features of clinical notes and GCN layers to encode the ICD code hierarchy boosted with code descriptions. 
Following the similar few-shot and zero-shot setting, Lu et al.~\cite{lu2020multi} improved the ZAGCNN model with knowledge aggregation from multiple graphs, i.e., the predefined hierarchy, the semantic similarity graph of label description, and the label co-occurrence graph.  
Meta-LMTC~\cite{wang2021meta} extends the ZAGCNN model by using optimization-based Model-Agnostic Meta-Learning (MAML) algorithm~\cite{finn2017model} and two sampling strategies (i.e., instance- and label-based) for meta optimization. 
Unlike the ZAGCNN and its extensions, Song et al.~\cite{song2020generalized} adopted the generalized zero-shot learning method for ICD coding. 
The authors proposed an Adversarial Generative Model (AGM) and utilized the Hierarchical Tree (HT) and code descriptions to generate code-specific features with the generative adversarial networks, as shown in Fig.~\ref{fig:AGM-HT}. 
The generated features are further used to fine-tune the medical coding model for zero-shot codes. 
Those methods mentioned above rely on external knowledge sources such as code hierarchy and descriptions (see more introduction about the usage of auxiliary information in Sec.~\ref{sec:auxiliary}, specifically Sec.~\ref{sec:code_description} for code descriptions and Sec.~\ref{sec:code_hierarchy} for code hierarchy).
\textcolor{black}{CoGraph~\cite{wang2021few} constructs a heterogeneous word-entity graph to represent clinical notes and performs graph contrastive learning on the constructed graph to improve the model's capability on few-shot prediction. 
Contrastive learning explores the intra-correlation of word-entity graphs via sampling and the inter-correlation of word-entity graphs via sequential modeling of graphs at different clinical stages. 
During the graph construction of the CoGraph model, Wikipedia acts as the source of external knowledge to obtain entity nodes.
Ji et al.~\cite{ji-MT-Hyper} showed that task-conditioned parameter generation with additional task information improves zero-shot diagnosis prediction. 
Auxiliary knowledge plays an essential role in few-shot and zero-shot medical coding. 
}

\subsubsection{Autoregressive Generative Decoders}

A recent pertaining and fine-tuning paradigm has been used for medical coding as introduced in Section \ref{sec:attention}.  However, in many cases, there is a significant gap between the goals of the downstream tasks and the pretraining goals. 
Moreover, specific fields require large amounts of the supervised corpus during fine-tuning.
A new fine-tuning paradigm based on the pretrained language model, prompt tuning that gives some best cues as the task-specific context for pretrained generative language models, has emerged to address these challenges. 
This approach has proven effective in few-shot tasks \cite{scao2021many, gao2020making}. 
In medical coding, Yang et al. \cite{yang2022knowledge} addressed the long-tail challenge using a prompt-tuning technique for label semantics, representing the first attempt to apply prompts to multiple label classification tasks. Specifically, the authors added a series of ICD code descriptions as the prompt and incorporated them early with clinical notes. To further improve the performance of the medical coding, the authors proposed a knowledge-enhanced longformer that injected three domain-specific knowledge (hierarchy, synonyms, and abbreviations) and utilized comparative learning for additional pre-training. 
In a follow-up study, Yang et al. \cite{yang2023multi} further tackled the long tail challenge in multi-label classification by converting it into an autoregressive generation task. The authors exploited a SOAP structure (i.e., subjective, objective, assessment, and plan) to generate free text diagnoses and procedures, which is medical logic used by physicians to note clinical documentation. They then translated the generated text's description to infer ICD codes with clinical vocabulary constraints, which solves the hallucination issues of generative models.
Prompt tuning with generative language models provides a novel solution for medical coding. 
It has benefits to utilize the knowledge from large language models. However, it also has some limitations. 
For example, the hallucinated generation can dampen the coding accuracy and require some engineering efforts on controlled generation, specifically when there is a shift in coding guidelines. 
Also, autoregressive generative models have the scaling issue when generating tokens, making them slower than non-autoregressive methods~\cite{yang2023multi}.

\subsubsection{Summary}

The problem setup and the principle of learning paradigms are the main motivators for choosing the decoder module.  
Neural attention decoders improve the fully connected layer-based decoder in the standard supervised learning setup to prioritize the representation learning on important information. 
The hierarchical decoders fit the hierarchical nature of medical codes. 
Multitask decoders aim to predict medical codes of multiple coding systems. 
Few-shot and zero-shot decoders solve the learning problem with rare or unseen medical codes.
And the autoregressive generative decoders, as an emerging approach, utilize the reasoning capacity of large autoregressive language models. 
\textcolor{black}{
The choice of decoder module in medical coding models should align with the problem's nuances and learning paradigms. Each type of decoder offers unique advantages, from enhanced attention mechanisms for information prioritization to specialized hierarchical structures, multitasking capabilities, adaptability to rare codes, and advanced autoregressive reasoning. 
Careful consideration of these factors is essential to develop effective and context-aware medical coding systems that can meet the diverse needs of healthcare professionals and patients.
}

\subsection{Usage of Auxiliary Information}
\label{sec:auxiliary}
Auxiliary information can be utilized to enhance representation learning and improve the performance of medical coding. 
This section introduces the usage of auxiliary information, including implicit information such as label information via randomly initialized embeddings and explicit information (or external data) such as Wikipedia articles, textual code descriptions, and code hierarchies. 
Implicit label information has been used by most previously introduced label attention-based models. 
The joint embedding model (LEAM)~\cite{wang2018joint} embeds labels and leverages the compatibility between word and label embeddings to calculate attention scores.
The following paragraphs review the methods that use external data explicitly. 
The external data can be applied to both encoders and decoders. 
When applied to encoders, external data enhance the representation learning of clinical texts.
The external information usually acts as the regularization for decoders when combining external data augmentation with the decoding process.

In addition to explicit usage of auxiliary information, data augmentation methods can also be applied to enrich the training data. 
Kim and Ganapathi~\cite{kim2021read} introduced a simple sentence permutation method to augment the training data three times and improve code prediction performance. 

\subsubsection{Wikipedia Articles}
Wikipedia articles explain medical diagnoses in detail and are used to enhance the deep learning model on clinical text understanding.
Prakash et al.~\cite{prakash2017condensed} resorted to Wikipedia as an external knowledge source.
Specifically, the authors used term search to find relevant articles to the diagnoses in clinical notes. They proposed C-MemNN with an iterative condensation of memory representations that utilize external knowledge sources from Wikipedia to enhance memory networks by preserving the hierarchical structure in the memory. 
KSI~\cite{bai2019improving} in Fig.~\ref{fig:KSI} uses element-wise multiplication and attention mechanism to fuse the knowledge from Wikipedia articles into clinical notes. 
There are 389 available Wikipedia pages when considering the first three digits of ICD-9 diagnosis codes. 
The KSI model defines the medical coding task as a classification problem of 344 ICD codes found in the code vocabulary of the used dataset.
Following the same setup, MCDA~\cite{wang2022novel}, a medical concept-driven attention model, aligns the clinical notes and Wikipedia articles in the latent topic space based on topic modeling. 
The joint embedding or alignment of Wikipedia and clinical notes introduces external knowledge sources to medical coding models. 
However, because some specific medical codes have no corresponding Wikipedia pages, the usage of KSI is only limited to coding three-digit ICD-9 codes, i.e., diagnostic category classification.
The absence of fine-grained coding may lead to the ineffectiveness of medical coding models in rare diagnoses or procedures.

\subsubsection{Code Description}
\label{sec:code_description}
\begin{figure}[htbp!]
\begin{center}
\begin{subfigure}[b]{0.5\textwidth}
	\includegraphics[height=4cm]{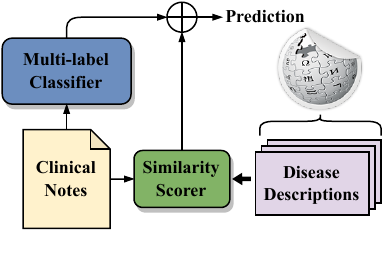}
	\caption{KSI~\cite{bai2019improving}}
	\label{fig:KSI}
\end{subfigure}
\qquad
\begin{subfigure}[b]{0.4\textwidth}
	\includegraphics[height=4cm]{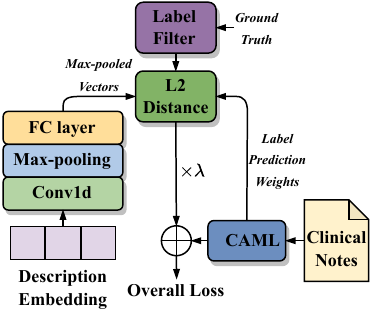}
	\caption{DR-CAML~\cite{mullenbach2018explainable}}
	\label{fig:DR-CAML}
\end{subfigure}
\caption{Illustrations of models that DR-CAML and KSI infuse external text features for regularization and feature augmentation. (a) KSI augments features via the multiplicative interaction between note features and embeddings of Wikipedia articles (b) DR-CAML infuses code description via regularization. }
\label{fig:feature_augmentation}
\end{center}
\end{figure}

The textual description of medical codes describes the exact meaning of codes and provides extra semantic information for abstract codes.
The embeddings of code description are denoted as $\mathbf{D}\in \mathbb{R}^{m\times d_t}$, where $m$ is the number of codes, and $d_t$ is the dimension of description embedding.
Several publications utilize the code description to enhance representation learning. 
DR-CAML~\cite{mullenbach2018explainable} as shown in Fig.~\ref{fig:DR-CAML} uses the word vectors of description as a regularization when optimizing the label-wise attention module. 
Similarly, CAIC~\cite{teng2020automatic} develops cross-textual attention to establish the connection between medical notes and ICD codes. 
GatedCNN-NCI~\cite{ji2021medical} builds fully connected interaction between notes and codes.
BiCapsNetLE~\cite{bao2021medical} uses embeddings of ICD descriptions to inject label information into the word embeddings of clinical notes and the features learned by capsule networks.
DLAC~\cite{feucht2021description} proposes a description-based label attention that computes the label attention matrix  with the description matrix and transformed hidden representation matrix as
 \begin{equation}
 	\mathbf{A}=\operatorname{Softmax}\left(\mathbf{H} \mathbf{U} \cdot \mathbf{D}^{\top}\right),
 \end{equation}
where $\mathbf{U}\in \mathbb{R}^{d_h\times d_t}$ is a transformation matrix that aligns the dimensions of the hidden representation and the description matrix. 
\textcolor{black}{A prompt-based fine-tuning model \cite{yang2022knowledge} adds a series of ICD code descriptions as the prompt to integrate code description and input notes for multi-label few-shot ICD coding.
}

\subsubsection{Code Hierarchy}
\label{sec:code_hierarchy} 
\begin{figure}[htbp!]
\begin{center}
\begin{subfigure}[b]{0.4\textwidth}
	\includegraphics[width=\linewidth]{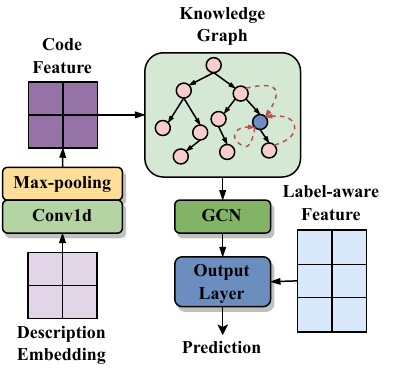}
	\caption{MSATT-KG~\cite{xie2019ehr}}
	\label{fig:MSATT-KG}
\end{subfigure}
\qquad
\begin{subfigure}[b]{0.4\textwidth}
	\includegraphics[width=\linewidth]{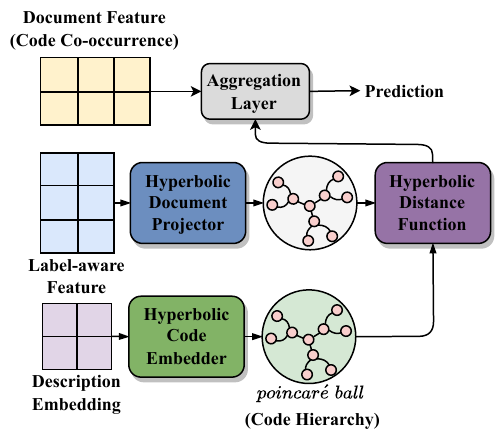}
	\caption{HyperCore~\cite{cao2020hypercore}}
	\label{fig:HyperCore}
\end{subfigure}
\caption{Illustrations of models that incorporate ICD hierarchy. (a) MSATT-KG uses GCN to encode the ICD code hierarchy. (b) HyperCore also uses GCN but in hyperbolic space.}
\label{fig:ICD_hierarchy}
\end{center}
\end{figure}

As introduced in Sec.~\ref{sec:decoder_hier}, hierarchical decoders use the code hierarchy. 
MSATT-KG~\cite{xie2019ehr} in Fig.~\ref{fig:MSATT-KG} infuses code hierarchy into document representation via structured knowledge graph (KG) propagation and label-dependent attention, where the code hierarchy is treated as a KG, and the graph convolutional network (GCN) is used to capture code relationships.
Similar to MSATT-KG, HyperCore~\cite{cao2020hypercore} also uses GCN to encode the code co-occurrence. 
Besides, it utilizes hyperbolic embedding and co-graph representation with code hierarchy, as shown in Fig.~\ref{fig:HyperCore}. 
HieNet~\cite{wang2022hienet} builds a bidirectional hierarchy passage encoder, consisting of a bidirectional passage retriever and a tree position encoder, to represent the code hierarchy with semantic and positional features. 
When classifying frequent codes with no significant hierarchical connections, Michalopoulos et al.~\cite{michalopoulos2022icdbigbird} built a co-occurrence graph of ICD codes with edge weights measured by normalized point-wise mutual information and applied graph convolutional networks to encode the ICD codes. 
\textcolor{black}{Lu et al. \cite{lu2024towards}  represented the ICD hierarchy as a super-tree and introduced tree editing distance \cite{zhang1989simple} to capture disease relationships at the code hierarchy.}
The hierarchical structure of the code system is a unique characteristic of medical coding, especially for predicting the complete code set. 
It is an exciting research direction that has the potential to improve coding performance and produce reliable and interpretable coding results.

\subsubsection{Medical Ontology}
\textcolor{black}{
We generally consider ontologies as TBox (i.e., concept-level knowledge that consists of logical statements constraining concepts) and ABox (data-level knowledge that consists of assertions over instances)~\citep{baader_horrocks_lutz_sattler_2017chap2}. A taxonomy (or hierarchy) can be reduced from the TBox of an ontology through a reasoning process called classification \cite[p.~34]{baader_horrocks_lutz_sattler_2017chap2}\footnote{The classification as a Description Logic reasoning process \cite{baader_horrocks_lutz_sattler_2017chap2} is different from its usage in machine learning.}. Also, taxonomy has a DAG (directed acyclic graph) structure.
Ontologies in the context of medical coding encompass a wide spectrum of structures, each with distinct characteristics and relationships; they are broadly referred to as terminologies or classification systems \cite{chap23HIbookCoiera}. 
}
In fact, ICD has the form of an ontology. 

Apart from that, the Unified Medical Language System~(UMLS) is a comprehensive collection of dictionaries and ontologies of in-domain concepts~\citep{lindberg1993unified,bodenreider2004unified}.
The UMLS is mainly based on three knowledge sources: Metathesaurus, Semantic Network, and SPECIALIST Lexicon and Lexical Tools.
Some studies explore the significance of ontology comprehension in designing effective deep-learning models for medical coding tasks.
MSMN~\cite{yuan2022code} extracts ICD code synonyms from the UMLS and proposes multiple synonym-matching networks to encode synonym information.
Dong et al.~\cite{dong2022ontology} leveraged the UMLS as an intermediary dictionary to extend the annotation vocabulary of rare diseases for their identification from clinical notes. 
Falis et al.~\cite{falis-etal-2022-horses} used UMLS (and their concept matching to ICD) to augment training data for medical coding.
The ability of a model to capture and utilize these intricate relationships can greatly impact its accuracy in code prediction. 
Consequently, understanding the specific ontology used in a medical coding task becomes crucial in selecting, adapting, or developing deep learning models.

While some ontologies, like the ICD, primarily rely on hierarchical relations without directional relationships, others feature more complex and nuanced relationships, such as ``causes'' and ``is-caused-by'' disease relations. 
These variations in ontology structures can have profound implications for the applicability and effectiveness of the deep learning approaches. 
\textcolor{black}{
The structures of different ontologies, such as the organization of concepts, relationships, and attributes, vary significantly~\citep{hadar2006variations}. 
Approaches optimized for purely hierarchical taxonomies may require adaptation, for example, when dealing with ontologies that incorporate causal relationships. 
Assume we have ontology A and ontology B and consider the representation of the relationship ``caused by'' between diseases and risk factors. In ontology A, this relationship might be straightforward, while in ontology B, it could involve nested attributes and additional complexities. For example, in ontology B, the relationship might be represented as ``Cardiovascular Disease'' is ``caused by'' ``Genetic Factors'' AND ``Environmental Factors''. Deep learning models optimized for simpler relationships might struggle to capture the nuances of complex relationships in ontology B. The variability in how relationships are modeled affects the model's ability to understand and predict based on different ontological structures.
}

\textcolor{black}{
In the context of ontological representation, particularly within the Web Ontology Language (OWL) under the EL profile~\cite{OWL_2_EL_profile}, we exemplify the SNOMED CT~\citep{clinical-finding-attributes} that employs existential restrictions with attributes to articulate intricate relationships, such as those involving attributes like ``Due to'' (e.g., Cataract of the eye due to diabetes mellitus (disorder)) which represent causal relations. 
Some complex relations can be represented as triples, e.g., (A, Due to, B), but other complex relations require more expressive representation, for example, which involves conjunctions and nested existential restrictions~\citep{dong2023ontology,chen2023contextual}. 
In this aforementioned example, ``Cataract of eye due to diabetes mellitus'' is equivalent to a conjunction of various concepts and attributes under a nested expression: 
\texttt{Cataract of eye due to diabetes mellitus} $\equiv$ \texttt{Disease} $\sqcap$ $\exists$ \texttt{RoleGroup}. ($\exists$ \texttt{FindingSite}. \texttt{Structure of lens of eye} $\sqcap$ $\exists$ \texttt{AssociatedMorphology}. \texttt{Abnormally opaque structure}) $\sqcap$ $\exists$ \texttt{RoleGroup}. $\exists$ \texttt{DueTo}. \texttt{Diabetes mellitus}\footnote{Attributes like \texttt{DueTo} and \texttt{AssociatedMorphology} are in UpperCamelCase form. The expression is in SNOMED CT version 2024-02-01 for the concept ID 43959009. For more examples, see SNOMED CT browser at \url{https://browser.ihtsdotools.org/}.}.
For complex relations that can be represented with triples, knowledge graph embeddings are essential~\cite{ji2022survey}. Addressing more logically complex relations, the use of OWL, particularly EL++ embeddings, becomes imperative. For instance, OWL2vec*~\citep{chen2021owl2vec} and EL++ geometrical embeddings~\citep{kulmanov2019embeddings,xiong2022faithful} exemplify this approach. Notably, despite the wealth of research, there appears to be a gap in exploring the embedding of OWL ontologies, such as in SNOMED CT, within the context of clinical coding. This observation suggests a promising avenue for future studies in this domain.
}

\subsubsection{Chart Data}

Patients' chart data (or structured data) that record the physiological conditions of a patient can be used to enhance the performance of code assignments. 
Multimodal machine learning methods use text and chart data to predict medical code.
Wang et al.~\cite{wang2016diagnosis} proposed a multi-label annotation model that inputs topic embeddings from patient notes and feature encodings from patients' chart data. 
The diagnosis code assignment module incorporates a disease correlation graph to capture the disease correlation.  
Xu et al.~\cite{xu2019multimodal} used texts including discharge summaries, radiology reports, nursing notes, and tabular data such as admission, lab events, and prescriptions. 
The authors developed an ensemble learning method with text CNN applied to text representation learning and a decision tree applied to transformed numerical features. 
The experimental results show that effective modeling of multimodal data can improve the model's robustness and accuracy. 
\textcolor{black}{
Liu et al. \cite{liu2022treeman} proposed a tree-enhanced multimodal attention network, TreeMAN, to capture the decisive information in structured medical data in EMRs. The authors first processed structured medical data into tabular data, then inputted tabular data into a trained decision tree to obtain tree-based features. Finally, the text representation and tree-based features are fused into a unified multimodal representation through an attention mechanism.
}

\subsubsection{Entities and Concepts}
The text mentions about medical codes in clinical notes contain rich world knowledge. 
Apart from code descriptions and Wikipedia articles, several works also utilize entities and concepts that abstract the expressions in clinical notes. 
These methods usually use existing clinical ontologies such as the Unified Medical Language System (UMLS).
Entity recognition and concept extraction aim to augment the text feature or provide additional supervision signals. 
Wiegreffe et al.~\cite{wiegreffe2019clinical} combined information extraction and medical coding by utilizing the cTAKES knowledge extraction system to extract concepts and reported a negative finding of document-level clinical coding. 
Falis et al.~\cite{falis-etal-2022-horses} extracted in-text UMLS entities (and the matched ICD entities) with SemEHR \cite{wu2018semehr} and MedCAT \cite{Kraljevic2021} to augment new coded training data with synonym and sibling code replacement, and reported improved results s few-shot and zero-shot coding.
Yuan et al.~\cite{yuan2022code} obtained code synonyms by aligning concepts in the UMLS. 
Inspired by the multi-head attention mechanism~\cite{vaswani2017attention}, the authors proposed multiple synonyms matching networks that take code synonyms as queries to match clinical texts and improve code prediction. 
\textcolor{black}{
Yang et al. \cite{yang2022knowledge} injected three domain-specific knowledge from UMLS, i.e., hierarchy, synonyms, and abbreviations, into a knowledge-enhanced longformer that mitigates data sparsity and improves model performance.
} 
\textcolor{black}{Li et al. \cite{li2023towards}  introduced medical knowledge from UMLS to construct entity-level text heterogeneous graphs. The usage of external knowledge improves individual notes' local context feature extraction. 
Ge et al. \cite{ge2023dkec} utilized medical guideline knowledge from the Regional Documents of Old Dominion EMS Alliance \footnote{
https://odemsa.net/} and medical datasets to construct a heterogeneous graph of medical entities. This graph captures entity relationships to compensate for data scarcity during model training.}

\subsection{Human-in-the-loop} 
\textcolor{black}{
In medical coding, the designed model and system need to meet the use scenarios of clinical coders, reduce the manual coding cost and enhance coding accuracy \cite{li2023nidn, li2023automatic}. ICD coding models with explanations can support decision-making better. 
In clinical practice, domain experts tend to trust the prediction results with reasonable explanations \cite{wood2022model}.
Thus, human-in-the-loop artificial intelligence that collaborates AI models with human beings becomes a useful paradigm to support the development of computer-assisted clinical coding (CAC). 
Various studies exploit human-computer interaction and are applied to augmented reality, brain-computer interface, and user customization. 
Combining human intelligence with deep learning models for medical coding requires much effort. 
For example, a successful human-computer interaction must empower the human coder to improve coding efficiency and accuracy. 
One quantitative study on human-computer interaction showed that the computer system should cater to the user’s diverse needs while ensuring efficient, effective, and safe interaction \cite{kivijarvi2023instrumental}. 
This section considers human efforts as a form of auxiliary information in automated coding models. 
It reviews the literature on medical coding methods relevant to human-in-the-loop learning systems, such as general computer-assisted clinical coding, active learning for annotation, explainability, and human evaluation.
}

\textcolor{black}{
\subsubsection{Computer-assisted Clinical Coding}
CAC is a continuously developing technology that can improve the accuracy and quality of clinical coding and relieve the pressure on clinical coding personnel by assigning diagnostic and procedure codes from EHRs to automate clinical coding. 
Campbell et al. \cite{campbell2020computer} reviewed and discussed CAC literature, and their findings indicated that CAC positively impacts coding quality and accuracy. 
Additionally, clinical coding personnel should view CAC as an opportunity rather than a threat. CAC transforms medical coding into a knowledge-based environment, and the current role of clinical coding professionals is transformed into clinical coding editors or analysts \cite{morsch2010computer}. However, the clinical coding editor still has the ultimate responsibility. They can reject any inappropriate clinical coding suggestions by CAC software and send a consultation letter to clinicians to clarify ambiguous or contradictory documents \cite{smith2010transitioning}. The automated clinical coding workflow must still follow the clinical coding principles and specifications. 
Using machine learning methods or computer-assisted clinical coding to extract practical information from EHRs and assign medical codes is the actual demand of each medical health organization. 
Biomedical-named entity recognition and linking (NER+L) is committed to extracting concepts from texts in EHRs. 
Searle et al. \cite{searle2019medcattrainer}  integrated MedCATTrainer with the biomedical NER+L model, which leverages active learning to improve the underlying NER+L model. 
Moreover, they provided researchers with configurable interfaces to define annotations specific to their research problems. This interface makes specific annotations for configurable use cases of previously identified and linked concepts. 
}

\textcolor{black}{
\subsubsection{Active Learning}
Deep learning models need many annotation examples for training, and domain experts need a high cost to annotate these data. 
Recent studies investigated the adoption of human-in-loop learning. 
For example, active learning lets human annotators focus on the most informative data samples. 
Thus, the cost of manual labeling can be reduced. 
Ferreira et al. \cite{ferreira2021active} employed the active learning method to select the sample with an enormous amount of information, significantly reducing manual annotation costs while maintaining the model's performance.  
Specifically, the authors studied two strategies for selecting samples: uncertainty metric and correlation. The uncertainty metric assesses how uncertain the model is for a particular instance, while the correlation measures the similarity between instances. 
}

\subsubsection{Explainability}
\textcolor{black}{Human-in-the-loop systems in healthcare aim to strike a balance between leveraging the capabilities of machine intelligence while maintaining the expertise and judgment of healthcare professionals. In such systems, healthcare providers can trust the correctness of the machine intelligence system's output~\cite{cutillo2020machine}, and the explainability allows them to understand the system's reasoning.
Teng et al. \cite{teng2020explainable} designed an explainable  G\_Coder model that uses two methods to verify its explainability. The first is to employ attention to extract keywords from clinical notes and display the correlation between medical labels and evidence. The second is to use doctors to judge the results of attention allocation quantitatively.
Oberste et al. \cite{oberste2022user} established a framework for characterizing user knowledge and prior knowledge included in explanations by reviewing knowledge-informed machine learning in healthcare. The authors encourage future research to improve the explainability of the model while considering user background and experience to stimulate more trust in clinical settings.
}

\textcolor{black}{
\subsubsection{Human Evaluation}
Human evaluation is vital in assessing medical coding systems. 
Kim et al. \cite{kim2022can} compared automated medical coding systems to human coders and introduced the Read, Attend, and Code (RAC) model, which performed on par with human coders. 
They hired two professional coders to assign ICD-9 codes to 508 patient discharge summaries, establishing a human coding baseline. 
The human coding baseline exceeded the machine learning baseline by a factor of 3.9 in micro-Jaccard similarity.
}

\section{Benchmarking and Real-World Usage}
\label{sec:benchmark}
This section introduces the data for benchmarking medical coding models and the evaluation metrics to evaluate the performance.

\subsection{Data}
The MIMIC database, including MIMIC-II~\cite{saeed2011multiparameter} and MIMIC-III~\cite{johnson2016mimic}, is currently the most popular data source for the experimental study of medical coding.
\textcolor{black}{
MIMIC-IV-Note~\cite{mimic-iv-note} - the latest development of the MIMIC database - contains deidentified free-text clinical notes collected in the U.S.A. and can act as a good testbed for the latest medical coding models.
}
\textcolor{black}{
Searle et al.~\cite{searle2020experimental} argued that the defacto gold-standard codes assigned in MIMIC-III had not undergone secondary validation and constituted a silver-standard dataset. 
}
The publicly available MIMIC database has promoted research on medical coding. 
The 2007 Computational Medicine Challenge organized a shared task on ICD coding with a dataset from a hospital in the US~\cite{pestian2007shared}.
Another recent dataset is CodiEsp in Spanish, which mainly provides manual codes with in-text explanations (or evidence) of 1,000 Spanish clinical notes, but also English translations of the notes and publications with ICD-10 codes. The dataset was used in the CodiEsp track in eHealth CLEF 2020 \cite{miranda2020overview}.

In parallel to the public databases, many studies have also been conducted with private in-house patient notes. 
For example, Zhang et al.~\cite{zhang2020bert} used de-identified medical notes with ICD-10 codes from a hospital in the USA.
Teng et al.~\cite{teng2020automatic} built a dataset of outpatient medical records with ICD-10 codes collected from a first-class hospital in China. 
Chen et al.~\cite{chen2020towards} also performed medical coding with datasets with 275,797 EMR documents from two medical departments of top hospitals in China and released a Chinese medical knowledge graph\footnote{\url{https://github.com/PaddlePaddle/Research/tree/master/KG/ACL2020_SignOrSymptom_Relationship}}.
Rios et al.~\cite{rios2021assigning} used a dataset of pathology reports with ICD-0-3 codes collected from the Kentucky Cancer Registry.
Liu et al.~\cite{liu2021effective} assigned ICD-10 codes to clinical documents of Dutch and French Datasets.
Hansen et al.~\cite{hansen2022assigning} utilized patients' medication history for diagnosis coding using the Danish register data.
Teng et al.~\cite{teng2022review} introduced several existing datasets for ICD coding.
Clinical notes from patient records have strict administrative regulations to protect patient privacy. 
However, more public de-identified data will help evaluate the generalizability of medical coding models.

\subsection{Evaluation Metrics}
Medical coding uses evaluation metrics of the multi-label multi-class classification problem to evaluate the predictive performance. 
Standard evaluation metrics such as the area under the receiver operating characteristic curve (AUC-ROC) and F1-score with two averaging strategies (i.e., micro and macro) and precision at $k$ (P@$k$) are used by most publications. 
Micro scores consider all labels jointly and give more weight to frequent labels. 
We summarize the medical coding performance of several representative methods on MIMIC-III full code datasets in Table~\ref{table:ICDfull}. 
The MIMIC-III top-50 dataset contains samples with the top-50 frequent codes.
The hierarchical nature of medical codes leads to the need for hierarchical evaluation. 
Instead of using a flat evaluation that treats each code independently, CoPHE~\cite{falis2021cophe} proposes a set of metrics that represent the depth of nodes in a hierarchy, allowing to quantify incorrect but related codes and preserve the counts in the upper layers to assess the issues of under- or over-prediction. 
Weak Hierarchical Confusion Matrix (WHCM) \cite{falis-etal-2022-horses} further adapts the confusion matrix to the document-level multi-label setting to track within-family or out-of-family confusion of codes based on the assumptions of a code ``family'' in the hierarchy (e.g., diagnosis codes that share the same first three digits of ICD-9).

\begin{table}[htbp]
\footnotesize
\setlength\tabcolsep{4pt} %
\begin{center}
\begin{tabular*}{0.9\linewidth}{@{\extracolsep{\fill}} lrr|rr|rr }
\toprule
\multicolumn{1}{l}{\multirow{2}{*}{Models}} & \multicolumn{2}{c}{AUC-ROC} & \multicolumn{2}{c}{F1} & \multicolumn{2}{c}{P@k}   \\
& \multicolumn{1}{c}{Macro}       & \multicolumn{1}{c|}{Micro}      & \multicolumn{1}{c}{Macro}      & \multicolumn{1}{c|}{Micro}      &   
\multicolumn{1}{c}{8}      & \multicolumn{1}{c}{15}     \\ 
\midrule
\multicolumn{1}{l}{{CNN}}~\cite{mullenbach2018explainable} & 80.6 & 96.9 & 4.2 & 41.9 & 58.1& 44.3\\
\multicolumn{1}{l}{{BiGRU}}~\cite{mullenbach2018explainable} & 82.2 & 97.1 & 3.8 & 41.7 & 58.5& 44.5\\
\multicolumn{1}{l}{{CAML}}~\cite{mullenbach2018explainable} & 89.5 & 98.6 & 8.8 & 53.9 & 70.9& 56.1\\
\multicolumn{1}{l}{{DR-CAML}}~\cite{mullenbach2018explainable} & 89.7  & 98.5 & 8.6 & 52.9& 69.0& 54.8\\ 
\multicolumn{1}{l}{{MultiResCNN}}~\cite{li2020multirescnn} & 91.0 & 98.6& 8.5& 55.2& 73.4& 58.4\\
\multicolumn{1}{l}{{MSAAT-KG}}~\cite{xie2019ehr} & 91.0 & 99.2& 9.0& 55.3& 72.8& 58.1\\
\multicolumn{1}{l}{{LAAT}}~\cite{vu2020label}	& 91.9 & 98.8& 9.9 & 57.5& 73.8& 59.1\\
\multicolumn{1}{l}{{JointLAAT}}~\cite{vu2020label}	& 92.1 & 98.8& 10.7& 57.5& 73.5& 59.0\\
\multicolumn{1}{l}{{HyperCore}}~\cite{cao2020hypercore} & 93.0 & 98.9 & 9.0 & 55.1& 72.2& 57.9\\
\multicolumn{1}{l}{{GatedCNN-NCI}}~\cite{ji2021medical} & 92.2 & 98.9 & 9.2 & 56.3 & 73.6& -\\
\multicolumn{1}{l}{{Fusion}}~\cite{luo2021fusion} & 91.5 & 98.7 & 8.3 & 55.4 & 73.6 & -\\
\multicolumn{1}{l}{{MARN}}~\cite{sun-marn} & 91.3 & 98.8 & 11.6 & 58.4& 75.4 & 60.2 \\
\multicolumn{1}{l}{{JLAN}}~\cite{li2021jlan}	 & 91.8 & 98.8 & 9.7 & 56.7 & 74.1& 57.9\\
\multicolumn{1}{l}{{ISD}}~\cite{zhou2021automatic} & 93.8 & 99.0 & 11.9 & 55.9  & 74.5& -\\
\multicolumn{1}{l}{{RAC}}~\cite{kim2021read} & 94.8 & 99.2 & 12.7 & 58.6 & 75.4& 60.1\\
\multicolumn{1}{l}{{MDBERT}}~\cite{zhang2022hierarchical} & 94.2 & 99.2 & 10.4 & 57.6 & 75.0& 59.6\\
\multicolumn{1}{l}{{MSMN}}~\cite{yuan2022code} & 95.0 & 99.2 & 10.3 & 58.4 & 75.2& 59.9\\
\multicolumn{1}{l}{{HieNet}}~\cite{wang2022hienet} & 93.3 & 99.2 & 9.3 & 56.6 & 78.3& 65.0\\
\bottomrule
\end{tabular*}
\end{center}
\captionsetup{justification=centering}
\caption{MIMIC-III-full~(ICD code) data set results~(in \%). ``-'' represents the corresponding results are not reported in the original papers.}
\label{table:ICDfull}
\end{table}

\subsection{Practice in Public Health}

Manual medical coding in practice is not perfect; e.g., the overall medium accuracy of coding in the UK was around 83\% with a large variance among studies (50-98\%) surveyed in \cite{burns2011} around 2012. Errors in manual coding may be due to errors or incompleteness in the patients' data, subjectivity in choosing diagnostic codes, lack of coding expertise, or data entry errors \cite{coiera2015chapter24}. There are usually backlogs (of months or over a year) of records to be coded \cite{alonso2020}. According to the survey in \cite{campbell2020computer}, a computer-assisted coding system can potentially help improve coding accuracy, quality, and efficiency; however, the challenges lie in the requirements in the transition from a manual process to a computer-assisted coding environment. Coders should be able to revise the codes suggested by the system and be involved in the system development process \cite{campbell2020computer}. There are also other challenges in linguistics (e.g., hypothetical contexts) and data formats (e.g., hand-written notes).  %
A recent, ongoing project on deploying an NLP system for clinical coding is CogStack for the Artificial Intelligence in Health and Care Award in England \cite{cogstack2021}. CogStack uses word embedding and concept embedding-based NLP sub-module MedCAT \cite{Kraljevic2021} to extract contextual entities with mapping to concepts or codes in UMLS, SNOMED, and ICD-10. We refer readers to \cite{dong-automated-clinical-coding} for a detailed comment of manual and AI-assisted clinical coding from the perspective of public health in the UK.

In China, ICD coding is crucial in medical statistics, medical evaluation, medical insurance, billing, and other fields \cite{diao2021automated}. Currently, the widely used coding rule is ICD-10. 
The relevant departments regularly update the coding rules and standards and prepare the extended version according to ICD-10, expanding it from 4 bits to 6 bits \cite{wang2020study}.  
However, the vast number of ICD codes and regional differences in the coding system lead to version disparities that adversely impact the coding quality \cite{gao2021multi}. 
Moreover, coders must master domain knowledge, coding rules, and medical terminology to complete basic coding tasks \cite{chen2017automatic}. 
In a Class III hospital in China, only two coders are typically assigned to code ICD for approximately 2000 inpatients daily \cite{jia2017hybrid}. 
Meanwhile, clinical experts have indicated that data loss and distortion are significant issues resulting from imprecise or insufficiently detailed descriptions in clinical notes, posing a major challenge to clinical coding practice. 
With the continuous development of deep learning technology and hardware, various deep neural network methods for Chinese corpus have been applied to the current ICD automatic coding model~\cite{jia2017hybrid,chen2017automatic,yang2018clinical,gao2021multi,yu2019automatic,zhou2020construction,cao2020clinical,zhao2023automated}. 
\textcolor{black}{
Clinical experts are optimistic about the potential and inevitability of integrating deep learning technology into the clinical practice of ICD coding, which could alleviate the burden on manual coders and improve coding accuracy \cite{chen2021automatic}. 
However, the deep learning model is challenging to play a decision-making role independently \cite{cao2020clinical}. 
Clinical coders prefer to believe in the model with solid interpretability and high accuracy \cite{wood2022model}. 
To address this, clinical experts recommend building an ICD coding assistant system that provides queries, recommendations, prompts, and other functionalities, integrating medical knowledge and rules into the clinical coding model based on deep learning \cite{zeng2019automatic}. 
In particular, because the subjective clinical records are not rigorous and complete, it may be advantageous to conduct learning and reasoning based on objective and factual data (such as microbiological events and subscriptions) \cite{liu2022treeman}.
}

\textcolor{black}{
There are more case studies in other countries. 
In Finland, the use of ICD-10 codes is a widely adopted practice, with most healthcare providers generating structured diagnosis codes as part of their day-to-day operations. Medical coding is essential for the purposes of monitoring the quality of clinical care, billing, insurance processing, and clinical research \cite{komulainen2012suomalainen}. In the Finnish context, most of the medical coding is carried out within EMR systems, with 100\% coverage reached in 2007, and 74\% of healthcare operators managing at least 90\% of their referral exchange electronically today \cite{reponen2021tieto}. The high degree of standardization and digitization of records has made it easy to develop interoperable automated medical coding systems in the Finnish healthcare domain. NLP-based medical coding has been explored as a way to identify unnoticed medical conditions, such as sleeplessness and anxiousness disorders, from clinical notes. The Finnish Institute for Health and Welfare (THL) calls for more widespread structuring of medical information, as well as for a systematic assessment of automated systems for coding and recording purposes, paving way for more widespread use of automated medical coding \cite{hypponen2014sahkoisen}.
}
\textcolor{black}{
In Thailand, Ponthongmak et al.~\cite{ponthongmak2023development} developed deep learning models for ICD-10 coding.
}
In Germany, the statutory health insurance billing for outpatient care is based on the German Uniform Assessment Standard (EBM). Oberste et al. \cite{oberste2022supporting} designed an ML system for EBM coding and achieved advanced predictive performance.
\textcolor{black}{
This review provides a general introduction to real-world applications. 
However, as a technical review, we could not cover every detailed aspect in the real world. 
We refer readers to the original publications for details.
When applying deep learning-based medical coding methods to country-specific scenarios, the data distribution, languages, and coding schemes changed. 
Previous performant models that achieved good accuracy on the English MIMIC-III dataset might not work well with new datasets. 
For example, \citet{ponthongmak2023development} showed PLM-ICD and CNN-PubMedBERT performed well in the Thai datasets and translation can enable the adaptation of the pretrained models to Thai-English clinical text.  
When moving to another country, the results might vary. 
This review serves as a good guideline for researchers and developers to develop their own deep learning models for medical coding. 
To comprehensively address the specific nuances of medical coding practices in different countries — for example, identifying the problems solved by country-specific studies, understanding their differences, and examining the empirical results obtained — a dedicated quantitative survey focusing on each country's context would be more suitable.
}

\subsection{Open Sources and Tools}
This section introduces useful resources, including open-source software, model implementations, pretrained language models, and medical ontologies.

Open-source codes facilitate reproducible research on clinical NLP algorithms.
We collect recent representative model implementations, packages, and software for references, as summarized in Table~\ref{tab:model_implementation}. 
The Python programming language and the Pytorch deep learning framework dominate the landscape of deep learning-based medical coding.

Aitziber et al.~\cite{atutxa2017machine} developed a machine learning-based extraction system called DTEncoding to automatically generate diagnostic terms~(DTs) from electronic health records.
SNOMED CT Browser\footnote{\url{https://browser.ihtsdotools.org/}} enables users to look up different SNOMED CT editions and clinical healthcare terminologies based on the SNOMED member countries.
The National Center for Health Statistics~(NCHS) developed and maintained the Mortality Medical Data System~(MMDS) to automatically provide classification, entry, and cause-of-death information on death certificates.
Clinical-Coder~\cite{cao2020clinical} is an online system that assigns ICD-10 codes to Chinese clinical notes.
It develops a Dilated Convolutional Attention network with N-gram Matching Mechanism (DACNM) for automated medical coding. 
Its dilated convolution is the same as that used in DCAN~\cite{ji2020dilated}. 
It further augments the system with an explicit n-gram matching to capture the explicit semantic information. 
A video demonstration is available~\footnote{Available at \url{https://youtu.be/U4TImTwEysE}}.
The Clinical Classifications Software~(CCS) is a maintained coding system projecting ICD-9-CM codes into coarse-grained CCS codes, which can be used as ontology and disease classification codes.
\textcolor{black}{
Recently, AnEMIC~\cite{juyong2022anemic} has been released typically for automatic ICD coding.
It is an open-source framework that enables error-reduced data preprocessing, model training, and evaluation for automatic ICD coding.
The current release includes multiple convolutional neural networks and transformer-based models.
}

\begin{table*}[htbp!]
\footnotesize
\begin{center}
\begin{tabular}{l | c | c| c  }
\toprule
{Model / Software} & {\makecell[c]{Programming\\Language}} & {Framework} & {URL}\\
\midrule
HA-GRU~\cite{baumel2018multi}	&	Python & TensorFlow  & \url{https://github.com/talbaumel/MIMIC}\\
CAML\cite{mullenbach2018explainable}	&	Python & Pytorch & \url{https://github.com/jamesmullenbach/caml-mimic} \\
KSI\cite{bai2019improving}	&	Python & Pytorch & \url{https://github.com/tiantiantu/KSI} \\
LAAT\cite{vu2020label}	&	Python & Pytorch & \url{https://github.com/aehrc/LAAT} \\
MultiResCNN\cite{li2020multirescnn}	&	Python & Pytorch & \href{https://github.com/foxlf823/Multi-Filter-Residual-Convolutional-Neural-Network}{https://github.com/foxlf823/MultiResCNN}\\
DCAN~\cite{ji2020dilated}	&	Python & Pytorch & \url{https://github.com/shaoxiongji/DCAN} \\
MT-RAM~\cite{sun2021multitask}	&	Python & Pytorch & \url{https://github.com/VRCMF/MT-RAM} \\
MARN~\cite{sun-marn}	&	Python & Pytorch & \url{https://github.com/VRCMF/MARN} \\
ISD~\cite{zhou2021automatic} 	&	Python & Pytorch & \url{https://github.com/tongzhou21/isd} \\
CMGE~\cite{wu2021counterfactual}	&	Python & Pytorch & \url{https://github.com/CKRE/CMGE} \\
MSMN~\cite{yuan2022code}	&	Python & Pytorch & \url{https://github.com/GanjinZero/ICD-MSMN} \\
\textcolor{black}{AnEMIC~\cite{juyong2022anemic}} & Python & Pytorch & \url{https://github.com/dalgu90/icd-coding-benchmark} \\
\bottomrule
\end{tabular}
\end{center}
\caption{A summary of open-source model implementations and packages}
\label{tab:model_implementation}
\end{table*}

Much literature~\cite{huang2019clinicalbert,ji2021does,alsentzer2019publicly} shows that models suffer from performance degeneration without considering the in-domain adaptation.
We collect the clinical pretrained language models and present the summary in Table~\ref{tab:pretrain_model}.
Most pretrained language models use text in the majority languages such as English and Spanish for self-supervised pretraining.
The research community should also pay more attention to the minority languages and learning algorithms in the low-resource setting.

\begin{table*}[htbp!]
\footnotesize
\begin{center}
\begin{tabular}{p{6cm} | c| c | c | c }
\toprule
\textbf{Model} & \textbf{Language} & \textbf{Architecture} & \textbf{Size} & \textbf{Pretraining Scheme}\\
\midrule
\href{https://huggingface.co/emilyalsentzer/Bio_ClinicalBERT}{emilyalsentzer/Bio\_ClinicalBERT} & English  & BERT & Base & Continued Pretraining \\
\href{https://huggingface.co/yikuan8/Clinical-Longformer}{yikuan8/Clinical-Longformer} & English & LongFormer & Base &  Continued Pretraining  \\
\href{https://huggingface.co/yikuan8/Clinical-BigBird}{yikuan8/Clinical-BigBird} & English & BigBird & Base &  Continued Pretraining  \\
\href{https://huggingface.co/Tsubasaz/clinical-pubmed-bert-base-512}{Tsubasaz/clinical-pubmed-bert-base-512} & English & BERT & Base & Continued Pretraining \\
\href{https://huggingface.co/llange/xlm-roberta-large-english-clinical}{llange/xlm-roberta-large-english-clinical}  & English & XLM-RoBERTa & Large & Continued Pretraining \\
\href{https://huggingface.co/PlanTL-GOB-ES/roberta-base-biomedical-clinical-es}{PlanTL-GOB-ES/roberta-base-biomedical-clinical-es} & Spanish & RoBERTa & Base & from scratch  \\
\href{https://huggingface.co/caracena/clinical-bert-base-spanish-wwm-uncased}{caracena/clinical-bert-base-spanish-wwm-uncased} & Spanish & BERT & Base & NA  \\
\href{https://huggingface.co/caracena/clinical-roberta-base-spanish-wwm-uncased}{caracena/clinical-roberta-base-spanish-wwm-uncased} & Spanish & RoBERTa & Base & NA \\
\bottomrule
\end{tabular}
\end{center}
\caption{A summary of clinical pretrained language models}%
\label{tab:pretrain_model}
\end{table*}

\section{Discussion and Future Directions}
\label{sec:discussion}

Clinical notes generated by clinicians contain rich information about patients' diagnoses and treatment procedures. 
Healthcare institutions digitized these clinical texts into EHRs and other structural medical and treatment histories of patients for clinical data management, health condition tracking, and automation.
This paper reviews deep neural network-based methods for automated medical coding under a unified framework. 
The advances in deep learning models have significantly improved predictive performance. 
\textcolor{black}{
However, the current trend of leaderboard-oriented research is also concerning. 
It is easy to fall into the pitfall of excessive neural architecture engineering by chasing the scores on the leaderboard of public benchmarks but missing other critical matters. 
}
For example, the widely used MIMIC-III dataset may be imperfectly labeled, or under-coded \cite{searle2020experimental}.
Moreover, it only reflects clinical coding practice in the US more than a decade ago (till 2012) and only contains clinical notes written in English.
This section summarizes the current research and discusses some critical issues that should be carefully considered when improving predictive performance.

\subsection{Summary and Discussion}
Our review proposes a unified framework for automated medical coding and categorizes building blocks under the unified framework. 
First, we introduce four main neural encoder modules, i.e., recurrent neural networks, convolutional neural networks, neural attention mechanism, and graph neural networks. 
Hierarchical encoders that utilize the hierarchical structure of the text are also discussed in detail. 
RNNs can capture the sequential dependency in text and are intensively used as the text encoder. 
Many models also use CNNs to extract local features, which is validated to be effective in medical coding with many labels. 
The neural attention mechanism, especially the self-attention-based Transformer network, suffers from quadratic complexity to the length of the input sequence. 
A recent study on hierarchical neural architecture with Transformer-based encoders shows that better construction of hierarchical text structure can improve the coding performance~\cite{zhang2022hierarchical}.
More efficient attention mechanisms such as Longformer~\cite{beltagy2020longformer} and Big Bird~\cite{zaheer2020big} for long sequences are also attracting much research attention while still very resource- and time-demanding compared to CNN models.

Most models stack neural layers to build deeper architectures. 
This review then introduces mechanisms to build deep neural networks. 
The most frequently adopted method is the residual connection initially proposed for image processing. 
The residual networks can avoid performance degradation of deep neural architectures and have been used in many CNN-based models.
However, the highway networks that use a gating mechanism to control the information flow of deep networks have not been used in building deep models for automated medical coding. 

Automated medical coding as a multi-label multi-class classification problem relies on powerful decoder modules to boost predictive performance. 
The most widely used decoder is the label attention mechanism that learns label-aware representations for medical code prediction.
The hierarchical nature of medical classification systems leads to the development of various hierarchical decoders.
Other advances, such as multitask learning and few-shot/zero-shot learning-based decoders, are also promising directions.

Finally, we review how to utilize auxiliary information to improve medical coding performance.
Auxiliary information includes Wikipedia articles, code descriptions, code hierarchy, chart data, entities, and concepts. 
Most methods use external text such as medical code descriptions and Wikipedia articles to enhance textual feature learning. 
Auxiliary information, such as code description, can also act as the regularization for model optimization. 
For example, the DR-CAML~\cite{mullenbach2018explainable} uses the embeddings of ICD code description as a regularizer to enhance the representation learning for rare codes.
The code hierarchy connects to hierarchical decoders closely. 
Chart data and medical imaging data facilitate multimodal learning. 
Entities and concepts enable knowledge-aware representation learning.  
External knowledge contributes to robust few-shot and zero-shot medical coding significantly. 
 
\subsection{Future Directions}
Deep learning has boosted the development of automated medical coding methods. 
Nevertheless, many challenges exist. 
This section points out some future directions as follows.

\paragraph{Long-term Dependency and Scalability}
It is challenging for neural encoders to capture long-term dependency, especially when clinical notes are extremely long documents.
Self-attention-based models that succeed in sentence understanding have scalability issues due to the complexity of self-attention.
Although some remedies attempt to make self-attention more efficient in the NLP community, few studies have been done in the context of medical coding.
Also, recent deep learning models are becoming increasingly large.
Future work should consider the scalability issue when dealing with long clinical documents and high-dimensional medical codes.

\paragraph{Clinical Relatedness}
Modern neural models can effectively learn textual features for given input texts. 
We can usually achieve satisfactory performance with a strong classifier and appropriate training.
However, whether the encoding model can capture the clinical relatedness between different text mentions for medical coding is still unclear.
Besides, human clinical coders refer to different data types when assigning codes. 
Multimodal deep learning methods are introduced to learn embeddings of multimodal data.
Multimodal alignment and fusion are critical components to capture clinical relatedness across different modalities.
Future work needs to deeply infuse clinical knowledge (e.g., knowledge graphs~\cite{chari2020directions,ji2022survey}) into the neural encoders, enhancing the model's capability to learn knowledge-aware features and the model's reasoning ability.

\paragraph{Class Imbalance and Hierarchical Decoding}
The medical coding tasks suffer from class imbalance with a long tail of rare diagnoses in the class distribution.
Nevertheless, current research considers less about the class imbalance issue, which should be addressed in future work.
In particular, more effective neural decoders would be required for robust medical coding. 
The code hierarchy as prior human knowledge sheds light on the imbalanced classes. 
However, how to enable global and local learning for the whole hierarchy and local branches in the hierarchy is a challenging future work when developing hierarchical decoding approaches. 
More importantly, enabling few-shot and zero-shot learning for rare and unseen codes without external knowledge is an unsolved problem.

\paragraph{Interpretability}
Existing models with a certain level of explanation are post-hoc studies, for example, by interpreting the predictions through the visualization of attention weights~\cite{dong2021explainable,feucht2021description}.
It is important to understand the model's prediction and prioritize features learned by the model.
For example, embedding external medical knowledge bases like the unified medical language system might be further utilized to learn rich knowledge-aware representation to help medical text understanding.
Knowledge-aware reasoning will need to be introduced to improve the interpretability of medical prediction. 
\textcolor{black}{
However, the medical coding model, mainly occupied by deep learning, is still largely a black box. 
Thus, further work should focus more on interpretability that can improve the transparency of neural medical coding models, explain and justify model predictions, and ensure accountability and adherence to privacy and ethical guidelines. 
}

\paragraph{Human-in-the-loop Systems}
Medical coding models are supposed to facilitate current workflows at hospitals. 
\textcolor{black}{
One important future research direction is to integrate the human-in-the-loop systems~\cite{zanzotto2019human} into medical coding, in which human experts can interact with the model training process and enhance the model performance \cite{holzinger2016interactive}.
} 
For example, the active learning paradigm can select informative samples for human annotators when preparing the training data. 
Hospitals usually provide coding guidelines for human coders. 
For example, one rule from the current guideline requires that hypertension with pregnancy should not be coded as hypertension.
Involving human experts can explicitly inject those coding rules into the labeled data and correct the model predictions if they fail to follow the coding guidelines.

\paragraph{Updated Guidelines and Data Shift}
Coding guidelines are usually updated frequently. 
The changes in guidelines should be considered in developing automated medical coding tools to facilitate the updated workflows at hospitals. 
As time goes by, clinical practice can change. 
For example, a new pandemic might lead to significant changes in the health systems.
Medical coding models should also be able to be robust to data shifts.
Considering the increasing nature of health records, incremental learning or lifelong learning~\cite{parisi2019continual} might also be studied. 
Multitask learning~\cite{zhang2021survey} that solves the medical coding with different coding schemes can also be further deployed. 
When new codes enter the standard classification system during the update of coding guidelines, the medical coding model should be adaptable to the zero-shot coding problem. 
Furthermore, medical coding models should also produce uncertainty-aware predictions when facing updated coding guidelines and schemes. 

\paragraph{Novel Encoder-decoder Architectures and Large Language Models}
\textcolor{black}{
While the existing work can be generalized to an encoder-
decoder architecture, sequence-to-sequence (seq2seq) models are less explored for medical coding by modeling text sequence input to code sequence output. 
}
Seq2seq models have been applied for multi-label classification to model the correlation among labels \cite{yang-etal-2018-sgm}, and more recently for entity recognition and linking from texts \cite{decao2020autoregressive}. 
\textcolor{black}{Atutxa et al. \cite{atutxa2018ixamed} performed a sequence-to-sequence benchmark for ICD-10 coding, published as working notes. However, the paper did not reveal details about how ICD codes are decoded in a sequence-to-sequence manner\footnote{\textcolor{black}{The decoder generates the output sequence one element at a time. At each step, it takes the previously generated element and the current state as input to predict the next element in the sequence.}}.}
\textcolor{black}{
Motivated by the seq2seq machine translation model, \citet{atutxa2019interpretable} continued the study on a real seq2seq model, in which the decoder process considers the previous output when decoding a new medical code.
As a future direction, novel decoding processes of the seq2seq method can be explored to better capture the dependency of medical codes.
}
Considering the prominent generation capabilities enabled by large pretrained language models~\cite{yang2022large} (e.g., encoder-decoder based T5 \cite{2020t5} and decoder-only BioGPT~\cite{luo2022biogpt}) and instruction-tuned models such as ChatGPT\footnote{\url{https://openai.com/blog/chatgpt}}, our initial version of this review argued that a potential direction is to generate codes from input documents and prompt-based learning that leverages crafted or learned templates (or prompts) for generation, as surveyed in \cite{liu2023pre}, may also be incorporated to decode medical codes or concepts in the large hierarchical coding space.
Recent advances, such as Yang et al.~\cite{yang2023multi}, investigated autoregressive generation with prompts. 
Utilizing the ability of large language models (LLMs), such as ChatGPT, for medical coding would be an interesting direction. 
\textcolor{black}{ 
\citet{falis2024gpt35} studied zero-shot prompting and data augmentation with GPT-3.5 and showed the limited capacity of GPT-3.5.
LLM-based methods hold the potentail for contextual understanding on medical text benefiting from the inherent language capabilities of such models.
However, they struggle with improving the coding accuracy, especically dealing codes without exisiting examples.
Moreover, this direction comes with challenges related to data quality, domain-specific terminology, fine-tuning, and ethical considerations. 
The integration of LLMs into medical coding represents a challenging path toward improving accuracy and efficiency in this critical healthcare task.
}

\paragraph{\textcolor{black}{Privacy and Security Concerns}}
\textcolor{black}{Medical texts often contain both identifying information (such as a patient's name or other identifying characteristics) as well as sensitive information (such as health state or intimate knowledge of their life). Thus, privacy and security concerns must always be addressed when processing, analyzing, and utilizing text-form data. When utilizing machine learning-based models, one must always remember that such models will likely retain patient-specific information unless training data has been thoroughly anonymized. This inherent memorization aspect makes the sharing of models between organizations difficult, as it is arduous to ensure that no patient-specific information can be reverse-engineered from a given model.}

\section{Conclusion}
\label{sec:conclusion} 

Recent years have witnessed increasing attention to the problem of automated medical coding.
This paper reviews automated medical coding from an in-depth perspective that unifies a great variety of existing deep learning-based models into an encoder-decoder framework without losing technical nuances in each specific type of model. 
Specifically, we discuss 1) neural encoders with recurrent and convolutional networks, neural attention mechanisms, and hierarchical encoders typically used for long clinical notes; 
2) mechanisms to build deep architectures, including simple stacking, embedding injection, and residual connection; 
3) decoder modules with linear layers, neural attention, hierarchical and multitask decoders; 
4) the usage of auxiliary information such as Wikipedia articles, code descriptions, and code hierarchy. 
Besides, we introduce data for medical coding, the evaluation of medical coding models, and real-world practice. 
We summarize the limitations and point out future research directions at the end of this review.

\section*{Acknowledgments}
This work was supported by the Research Council of Finland (Flagship programme: Finnish Center for Artificial Intelligence FCAI, and grants 336033, 352986, 358246) and EU (H2020 grant 101016775 and NextGenerationEU). H. Dong is supported by Health Data Research UK National Phenomics
and Text Analytics Implementation Projects and EPSRC project (EP/V050869/1).
The authors acknowledge Matúš Falis for his helpful comments.

\bibliographystyle{ACM-Reference-Format}
\bibliography{references}

\end{document}